\documentclass[12pt]{spieman}  
\usepackage{amsmath,amsfonts,amssymb}
\usepackage{graphicx}
\usepackage{setspace}
\usepackage{tocloft}
\usepackage{algorithm}
\usepackage{algorithmic}
\usepackage{color}

\makeatletter
\let\c@lofdepth\relax
\let\c@lotdepth\relax
\makeatother
\usepackage{subfigure}

\title{Unsupervised convolutional neural network fusion approach for change detection in remote sensing images}

\author[a,*]{Weidong Yan}
\author[a]{Pei Yan}
\author[a]{Li Cao}
\affil[a]{School of Mathematics and Statistics, Northwestern Polytechnical University, Xi'an 710129}

\cftpagenumbersoff{figure}
\cftpagenumbersoff{table} 
\begin{document} 
\maketitle

\begin{abstract}

With the rapid development of deep learning, a variety of change detection methods based on deep learning have emerged in recent years. However, these methods usually require a large number of training samples to train the network model, so it is very expensive. In this paper, we introduce a completely unsupervised shallow convolutional neural network (USCNN) fusion approach for change detection. Firstly, the bi-temporal images are transformed into different feature spaces by using convolution kernels of different sizes to extract multi-scale information of the images. Secondly, the output features of bi-temporal images at the same convolution kernels are subtracted to obtain the corresponding difference images, and the difference feature images at the same scale are fused into one feature image by using $1\times 1$ convolution layer. Finally, the output features of different scales are concatenated and a $1\times 1$ convolution layer is used to fuse the multi-scale information of the image. The model parameters are obtained by a redesigned sparse function. Our model has three features: the entire training process is conducted in an unsupervised manner, the network architecture is shallow, and the objective function is sparse. Thus, it can be seen as a kind of lightweight network model. Experimental results on four real remote sensing datasets indicate the feasibility and effectiveness of the proposed approach.

\end{abstract}


{\noindent \footnotesize\textbf{*}Weidong Yan,  \linkable{yanweidong@nwpu.edu.cn} }

\begin{spacing}{2}   

\section{Introduction}\label{sect1}  

Remote sensing change detection aims to identify the changed and unchanged areas by comparing two or more remote sensing images acquired at different times over the same area \cite{1}. It has been widely used in many fields, such as agricultural surveys \cite{2}, urban planning \cite{3}, and disaster assessment \cite{4}. 

\par At present, most of the existing methods concentrate on generating a well-performing difference map with the reason that the subsequent analysis of the difference map and the final detection results to a large extent depend on the quality of the difference map. Among some research works, ratio operator and its variants, such as log-ratio operator \cite{5}, mean-ratio operator \cite{6,7}, frequently used to generate the difference maps. In particular, since the log ratio operator and the mean ratio operator have a good ability to reduce speckle noise, they are often used on pre-detection stage to produce training samples with labels in some supervised methods \cite{8,9}. As for the unsupervised methods, studies on the relevant work attract researchers’ attention as well. Moreover, it has been proven that many change detection methods available based on combination of these operators achieved better performance than using only one operator. Ma et al.\cite{10} proposed a novel method based on wavelet fusion, which generates a difference map via utilizing complementary information from log-ratio and mean-ratio images, and then choose weight averaging and minimum standard deviation as fusion rules to obtain the change map. Hou et al.\cite{11} employed a fused difference map of Gauss-log-ratio and log-ratio operators along with nonsubsampled contourlet transform (NSCT) and compressed projection strategy to detect changes, making it easier to separate change areas from the background information, while being immune to speckle noise. Yan et al.\cite{12} proposed a method based on frequency difference and a modified fuzzy c-means clustering for change detection. In this method, two images at different phases were both decomposed by wavelet transform and the log-ratio and log-mean-ratio (LMR) images are also constructed respectively according to the corresponding wavelet coefficients. Yan et al.\cite{13}  proposed a change detection method based on coupled distance metric learning (CDML).The method uses log-ratio operator as pre detection method to select training samples, and tries to learn a pair of mapping matrices by using the information of unchanged pixels and changing pixels to improve the contrast between changed pixels and unchanged pixels.

\par In recent years, influenced by the excellent performance of deep learning in image classification \cite{14,15,16}, researchers have been inclined to adopt methods based on deep learning to solve the problem of change detection for remote sensing. Liu et al.\cite{17}established a deep neural network using stacked Restricted Boltzmann Machines to analyze the difference map and recognize the changed and unchanged pixels, where the training process of network model relatively fully reflect the real changed features and trends. Gong et al.\cite{18} proposed a change detection method by training a deep neural network to produce the binary map directly from two images, so as to the negative effect on the results as much as possible. Since wavelet has a good performance in acquiring the local, multi-scale and other crucial information of the images, it gradually plays an increasingly irreplaceable role in dealing with the image-related problems including Image recognition, computer vision and so on. Gao et al.\cite{8} utilized Gabor wavelets and FCM to select samples and trained the PCANet to classify pixels features with the best results delivered over three real SAR image data sets compared with  four other approaches. Different from the conventional methods of obtaining training samples with label information,  Du et al.\cite{19} utilized CVA  to conduct the pre-detection process, both deep network and slow feature analysis theory combined to accomplish the change detection to validate robustness and effectiveness of the proposed algorithm.  Gao et al.\cite{20} incorporate wavelet transform into convolutional neural network in expectation of improving the detection performance. It is worth noting that the above methods \cite{18,19,20} all must go through a necessary pre-detection procedure aiming at choosing samples with higher accuracy. Thus, the final detection results are largely dependent upon the pre-detection results. In addition, network models based on deep learning are generally built with more layers, where shallower layers are able to learn a part of simple and specific feature representations. Although a multi-layer network outperforms in the classification of changed and unchanged areas by learning distributed representations from the two original images, as the number of layers of neural network deepens, the complexity of the whole network increases and training also becomes very time-consuming.

\par
\begin{figure}[htbp]
	\centering
	\includegraphics[width=14cm,height=7.0cm]{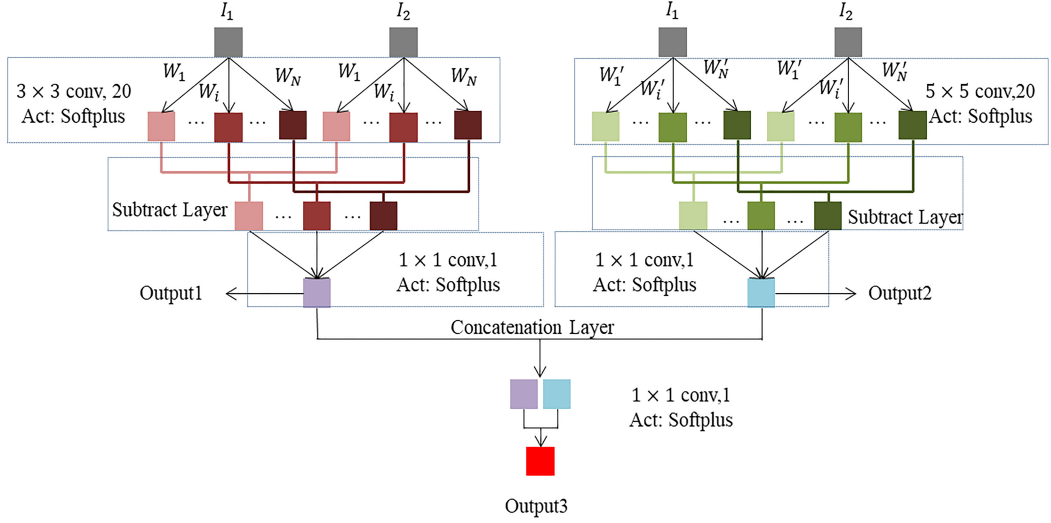}
	\caption{Structure of USCNN.}
	\label{figure1}
\end{figure}

 To make the network model lightweight, in this paper, a novel method based on Unsupervised Shallow Convolutional Neural Network (USCNN) fusion approach is presented. The framework of USCNN is shown in Fig. \ref{figure1}. The proposed method introduces the idea of the traditional log-ratio and mean-ratio operator to convolutional neural network interestingly, where the role of the mean operator is being a special convolution kernel and to effectively reduce speckle noise of the two original images. However, the parameters of the traditional mean operator in the filter have been fixed, so it cannot adaptively learn these parameters from different images which may limit the accuracy of the final results. To this end, we extend the idea of a combination of log-ratio operator and mean-ratio operator to the convolutional neural network, which can easily and adaptively learn filtering parameters without other auxiliary information. A point that needs to be noted here is that, in our model, we use two kinds of convolution kernels of different sizes ($3\times 3$ and $5\times 5$ in the model) to implement multi-scale information extraction of images. Our main contributions are threefold.

1)  Owing to error gradients certainly decay exponentially with depth of the model, we propose a completely unsupervised shallow convolutional neural network fusion approach for change detection. In the method, we design an objective function with sparse properties to train the network in an unsupervised fashion.

2)  In order to make full use of the multi-scale information of the image to suppress speckle noise, two filtering kernels of different sizes are used to gain more multi-scale information.

3)	To drop the interference of the multiplicative noise of SAR images, we integrate log-ratio operator and mean-ratio operator ideas into a convolutional neural network and our model can be feasibly embedded in models constructed by other operators.

\par The rest of the paper is organized as follows. In Section \ref{sect2}, we describe the motivation and details of the model. In Section \ref{sect3}, the parameter analysis and the experiment results are presented. Finally, Section \ref{sect4} concludes the paper.

\section{Methodology}\label{sect2}

\subsection{Motivation}\label{sunsect2.1}

\par Change detection algorithms on remote sensing images usually degrade the real performance owing to the presence of speckle noise. By contrast, Ban\cite{21} presented the traditional difference operator can often achieve great effect to some extent on optical remote sensing images. Bovolo and Bruzzone\cite{5} produced the Log-ratio operator can transform multiplicative noise into additive noise, so it usually performs better than difference operator in SAR images change detection. Let $I_1$ and $I_2$ be the two temporal images acquired at the same area and $DI$ be the difference map, so the log-ratio operator can be expressed as follows:

\begin{equation}\label{1}
DI=|\log I_1 -\log I_2 |
\end{equation}

\par
\begin{figure}[htbp]
	\centering
	\includegraphics[width=14cm,height=4.8cm]{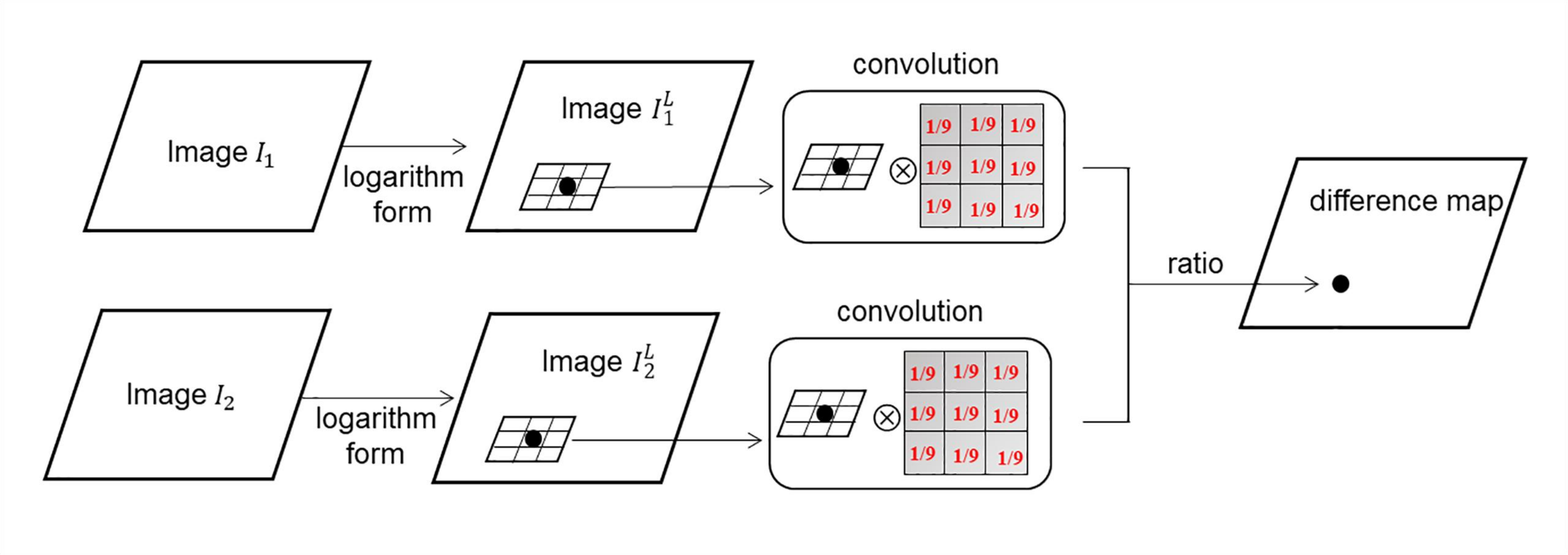}
	\caption{Flow chart of LMR operator.}
	\label{figure2}
\end{figure}

 To further suppress the influence of noise, the mean operator is introduced into the logarithm operator \cite{12}. Based on the log-ratio operator, the neighborhood patch is taken for each pixel of the original image, and the average value of the patch is calculated to replace the central pixel value. The flow chart of the LMR operator is shown in Fig. \ref{figure2}, and the calculation process can be expressed as follows:

\begin{equation}\label{2}
I^L_1=\log I_1,\quad I^L_2=\log I_2 
\end{equation} 

\begin{equation}\label{3}
DI(i,j)=|\frac{\mu_1(i,j)}{\mu_2(i,j)}|
\end{equation} 

\noindent where $I^L_1$ and $I^L_2$ denote new images after taking the logarithm operation on the two images $I_1$ and $I_2$, respectively. $\mu_1(i,j)$ and $\mu_2(i,j)$ denote the mean values of the neighborhood patch centered on pixel $(i,j)$ in $I^L_1$ and $I^L_2$, respectively.

\begin{figure}[htbp]
	\centering
	
	\subfigure[]{
		\centering
		\includegraphics[width=0.3\linewidth]{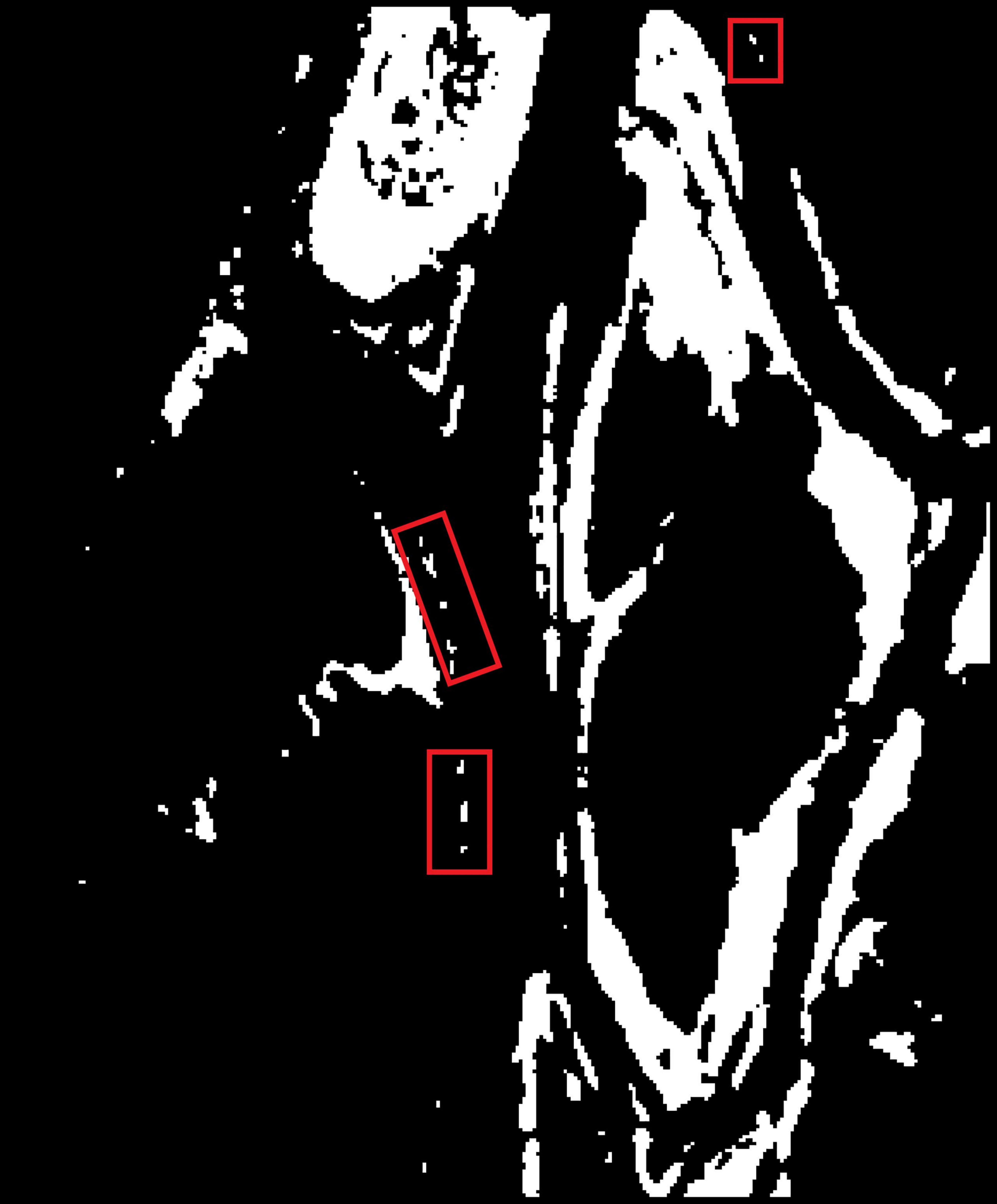}
	}
	\subfigure[]{
		\centering
		\includegraphics[width=0.3\linewidth]{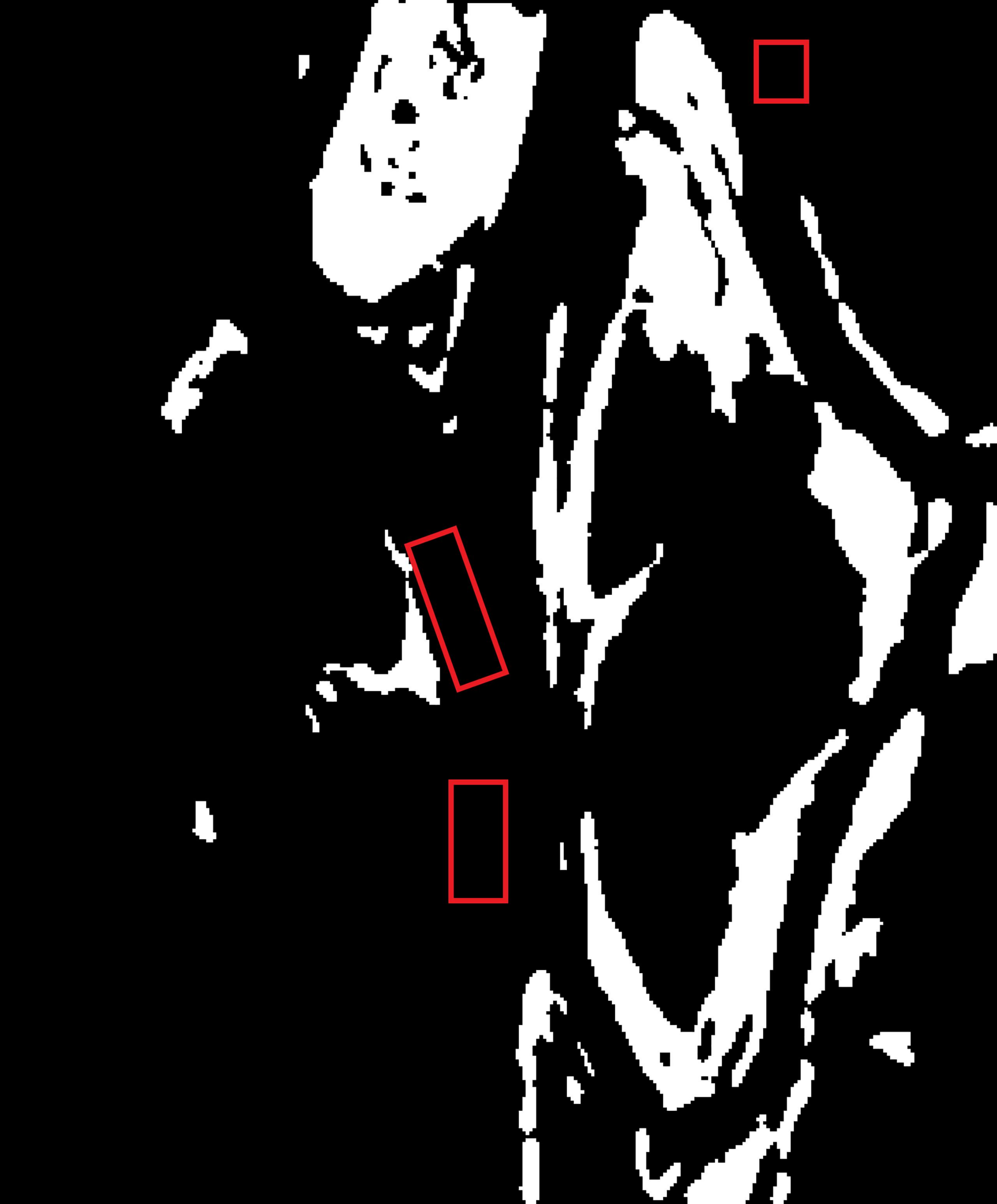}
	}
	\subfigure[]{
		\centering
		\includegraphics[width=0.3\linewidth]{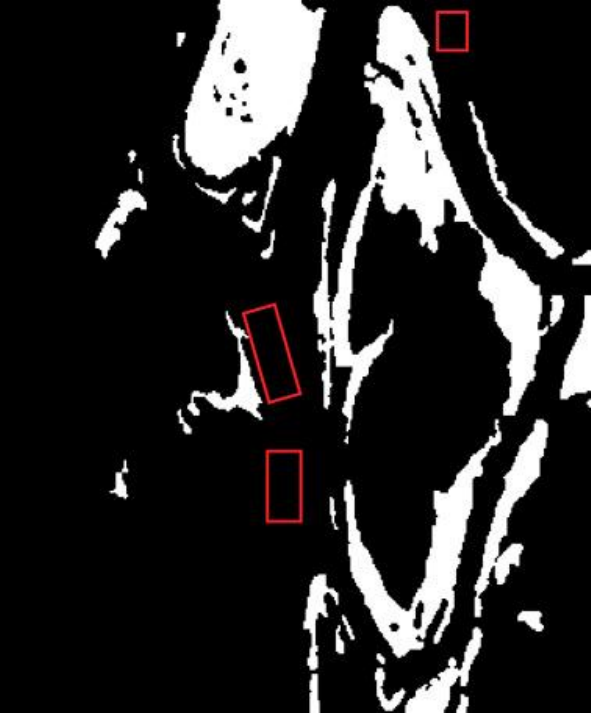}
	}
	\caption{ Binary maps. (a) LMR. (b) The proposed method. (c) Reference map.}
	\label{figure3}
	
\end{figure}

\par As shown in Fig.\ref{figure2}, The mean operation in the LMR operator is equivalent to the pixel-by-pixel convolution of the original image using a fixed convolution kernel. However, the convolution kernel with fixed parameters cannot adaptively learn these parameters from different images, so its performance is limited. Fig.\ref{figure3} (a) and (b) show the binary maps obtained by LMR operator and USCNN using k-Means \cite{22} clustering, and the ground truth is shown in Fig.\ref{figure3} (c). It can be observed that there exist many isolated noises in the red box located at the binary map which is achieved by the LMR operator. In fact, due to the fixed parameters limitation of the mean filtering kernel in LMR operator, some unchanged pixels are incorrectly classified as changed pixels. To address the issue, we present a new method to adaptively learn the filtering kernel parameters, which can effectively improve the accuracy of change detection. 

\subsection{The proposed method}\label{sunsect2.2}

\par As mentioned before, the mean-ratio operator can be regarded as a special convolution operation on the images. However, the mean operator limits the adaptive learning of the so-called convolution kernel parameters. The loss of adaptability is generally accompanied by an inadequate feature learning processes, which ultimately affect the interpretation of SAR images. To compensate for this deficiency, we advance an innovative algorithm, which is an unsupervised shallow convolutional neural network that can adaptively learn the convolution kernel parameters and greatly enhance the training speed of network.The structure of the USCNN is shown in Fig.\ref{figure1}.

\par USCNN is a symmetric shallow network with two branches. Our goal is to accomplish surprising results using multi-scale feature fusion. In detail, a convolution layer is also a feature extraction layer, to acquire richer information from two different scales, we set the convolution kernel sizes of a symmetric branch to $3\times 3$ and $5\times5$, respectively. To generate the related difference map pixel by pixel, we share the same weights $W_i$ or $W_{i}^{'}$ between the two inputs $I_1$ and $I_2$ of each branch. After that, the feature maps obtained using the convolution kernel with the same weights are subtracted at the pixel level to highlight the changed information. The calculation expression is given as follows:

\begin{equation}\label{4}
S_i=g_1 \circ (W_i \otimes I_1+b_i)- g_1 \circ (W_i \otimes I_2+b_i),i=1,2,...,N
\end{equation} 

\begin{equation}\label{5}
S_{i}^{'}=g_{1}^{'} \circ (W_{i}^{'} \otimes I_1+b_{i}^{'})- g_{1}^{'} \circ (W_{i}^{'} \otimes I_2+b_{i}^{'}),i=1,2,...,N
\end{equation} 

\noindent where $S_i$ and $S_{i}^{'}$ represent the $i$-th feature map of the subtract layer in the two branches, respectively. $g_1$ and $g_{1}^{'}$ represent the activate functions of the convolutional layers with $3\times 3$ and $5\times 5$ kernels, and in this paper we choose softplus \cite{23} as activation function. $W_i$ and $W_{i}^{'}$ represent the $3\times 3$ and $5\times 5$ kernels, respectively. $\otimes$ denotes the convolution operation. $b_i$ and $b_{i}^{'}$ represent the bias of the convolutional layers with $3\times 3$ and $5\times 5$ kernels. $N$ is the number of  convolutional kernels, and in this paper we set $N=20$.

\par After getting the feature maps $S_i$ and $S_{i}^{'}$, the convolutional layers with $1\times 1$ kernels \cite{24} are adopted to fuse the information from two output channels, and each branch gets a feature map. The calculation process can be expressed as follows:

\begin{equation}\label{6}
C=g_2 \circ (W \otimes S+b)
\end{equation} 

\begin{equation}\label{7}
C^{'}=g_{2}^{'} \circ (W^{'} \otimes S^{'}+b^{'})
\end{equation}

\noindent where $C$ and $C^{'}$ denote the final feature maps from two branches, respectively. $g_2$ and $g_{2}^{'}$ represent the activation functions. $W$ and $W^{'}$ denote the weights of $1\times 1$ convolutional kernels. $S=\lbrace S_i,i=1,...,N\rbrace$ and $S^{'}=\lbrace S_{i}^{'},i=1,...,N \rbrace$ represent the outputs of the upper layers of the two branches, respectively. $b$ and $b^{'}$ are bias.

\par A well-performed difference map is beneficial for the successive cluster analysis. In general, change maps are sparse. In an ideal state, for a good difference map, the pixel values of the unchanged area should be infinitely close to zero, and the pixel values of the changed area should be far from zero. Based on this fact, we introduce the sparse function into our objective function. It is defined as:

\begin{equation}\label{9}
f_1=\frac{1}{r \times c} \sum_{i=1}^{r}\sum_{j=1}^{c}|C(i,j)|
\end{equation}

\begin{equation}\label{10}
f_2=\frac{1}{r \times c} \sum_{i=1}^{r}\sum_{j=1}^{c}|C^{'}(i,j)|
\end{equation}

\noindent where $r \times c$ denote the size of two temporal images $I_1$ and $I_2$. $(i,j)$ represents the spatial position of image pixels.

To fuse the information of the two branches, we concatenate the two feature maps together and use a $1\times 1$ convolutional kernel to fuse the information from two branches. The calculation process can be expressed as follows:

\begin{equation}\label{8}
M=g_{3} \circ (\widetilde{W} \otimes (C \oplus C^{'})+\widetilde{b})
\end{equation}

\noindent where $M$ denote the final fusion result. $g_3$ represents the activation function. $\widetilde{W}$ and $\widetilde{b}$ represent the weight and bias, respectively. $\oplus$ indicates the concatenate operator.

\par The reason we use the sparse functions $f_1$ and $f_2$ is to make the pixel values in unchanged area of $C$ and $C^{'}$ close to zero. Our goal is to find a state in the training process so that the pixel difference between the changed area and the unchanged area can be perfectly separated. However, it is difficult to fulfill this purpose based only on $f_1$ and $f_2$, because the actual training process is harder to control than expected, and it is easy to make all the pixels of two feature maps go to zero. Based on this, we introduce a term for the fused output into the objective function, as follows:

\begin{equation}\label{11}
f_3=\frac{1}{r \times c} \sum_{i=1}^{r}\sum_{j=1}^{c}|M(i,j)|
\end{equation}

\par Clearly, $f_3$ and $f_1$ or $f_2$ are similarly defined, but they play different roles in the overall objective function(the expression will given later). $f_1$ and $f_2$ have the same goal, which tries to make the output of the two branches more sparse, particularly performing well in reducing isolated noise points in the result images. However, $f_3$ has the opposite goal that is trying to keep the pixels of the fusion result map of two branches away from 0, which not only prevents the two branches from excessively weakening the difference of the changed pixels but also improves the ability to distinguish between the changed and the unchanged pixels.

\par The overall objective function is defined as follows:

\begin{equation}\label{12}
L=f_1+f_2-k \cdot f_3, k>0
\end{equation}

\noindent where $k$ is a super parameter, which will be discussed in  Section \ref{sect3}.

\section{Experiments}
\label{sect3}

To validate the effectiveness of the method, we tested our method on four remote sensing datasets. In this section, we analyze the influence of the parameters and demonstrate the competitive performance on four real datasets. In section \ref{subsect3.1}, we first introduce the test datasets and change detection evaluation criteria. In section \ref{subsect3.2}, we analyze the effect of the parameter $k$. Finally, in section \ref{subsect3.3} , we compare it with four closely related change detection methods.

\subsection{Datasets description and evaluation criteria}\label{subsect3.1}
\begin{figure}[htbp]
	\centering
	
	\subfigure[]{
		\centering
		\includegraphics[width=0.3\linewidth]{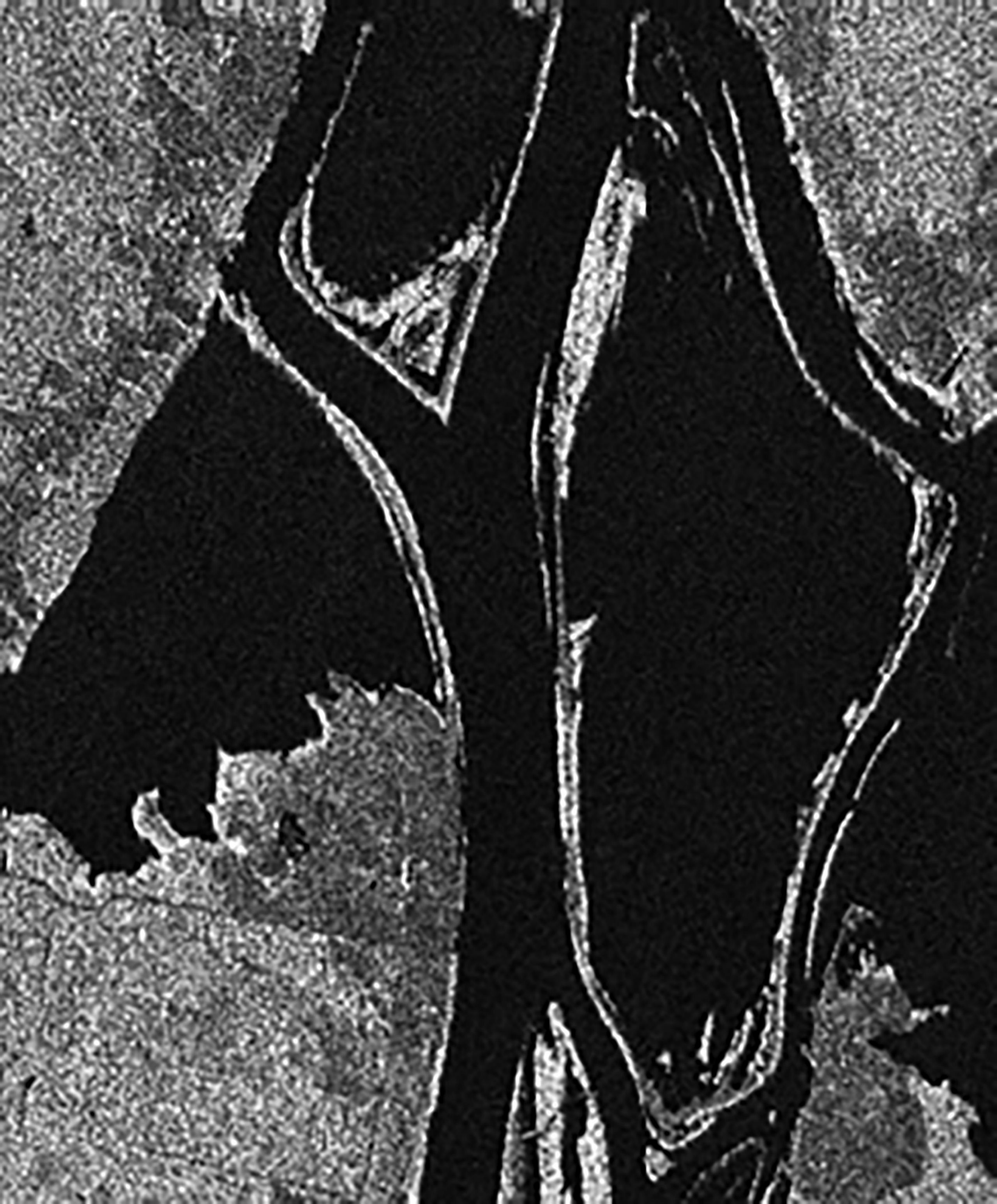}
	}
	\subfigure[]{
		\centering
		\includegraphics[width=0.3\linewidth]{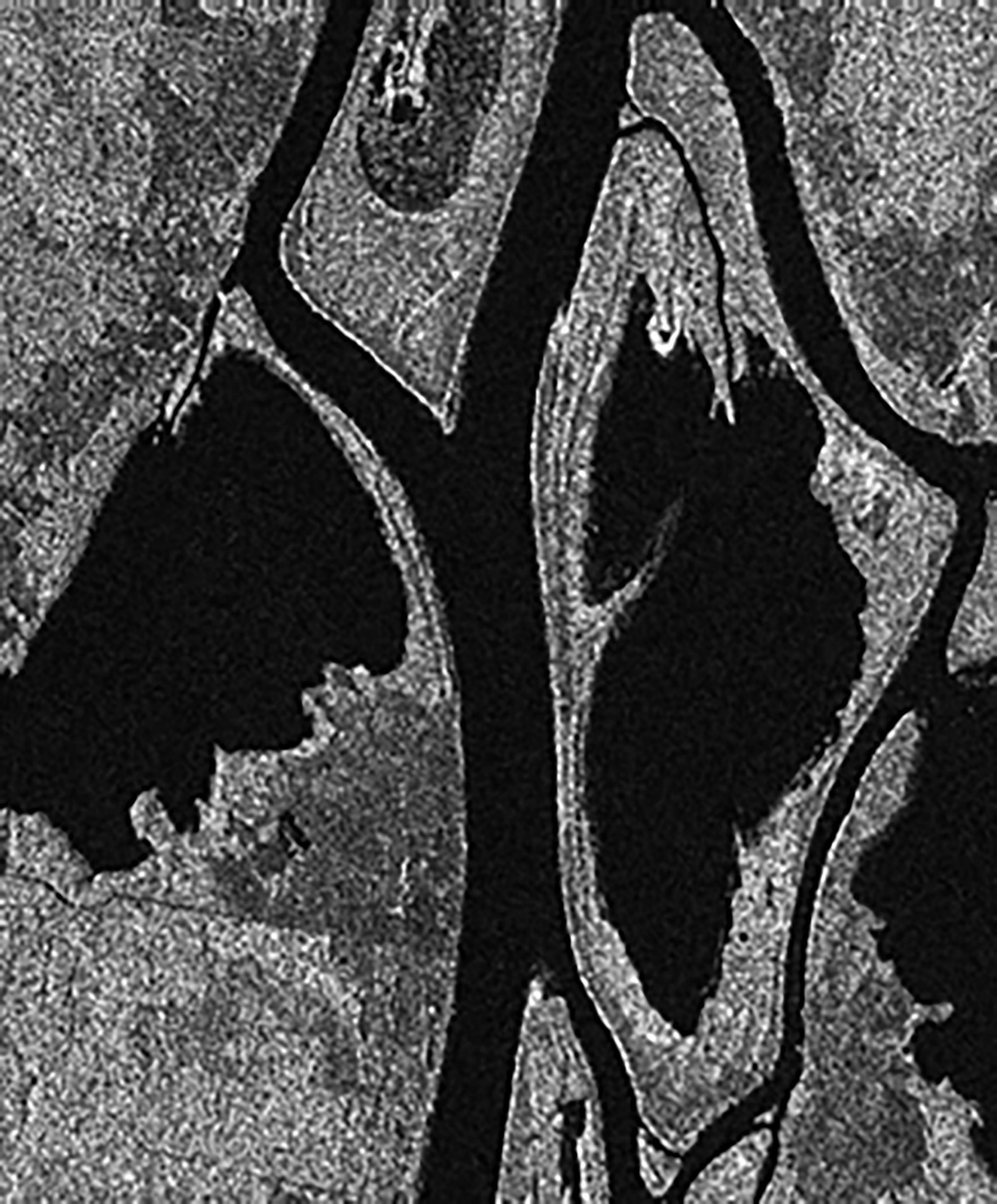}
	}
	\subfigure[]{
		\centering
		\includegraphics[width=0.3\linewidth]{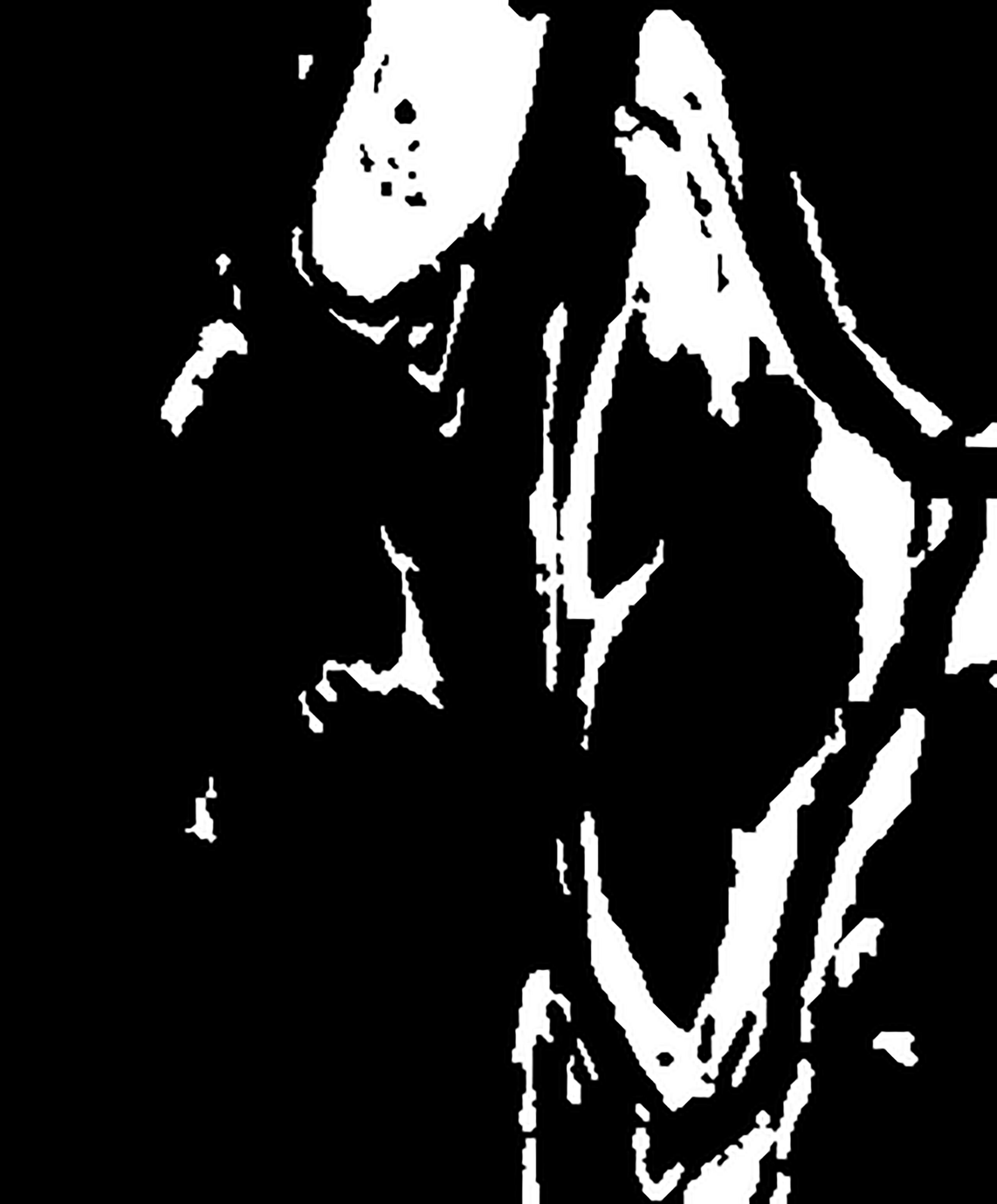}
	}
	\caption{ Images relating to the Ottawa: (a) Image acquired in September 1995. (b) Image acquired in July 1996. (c) Ground truth.}
	\label{figure7}
	
\end{figure}

The first dataset (shown in Fig.\ref{figure7}(a) and (b)) consists of bi-temporal SAR images, acquired by the RADARSAT at the area of Ottawa in May 1997 and August 1997, respectively. In addition, The ground truth reflecting regional changes is shown in Fig.\ref{figure7} (c). The size of the three images is $290 \times 350$ pixels. 

\par
\begin{figure}[htbp]
	\centering
	
	\subfigure[]{
		\centering
		\includegraphics[width=0.3\linewidth]{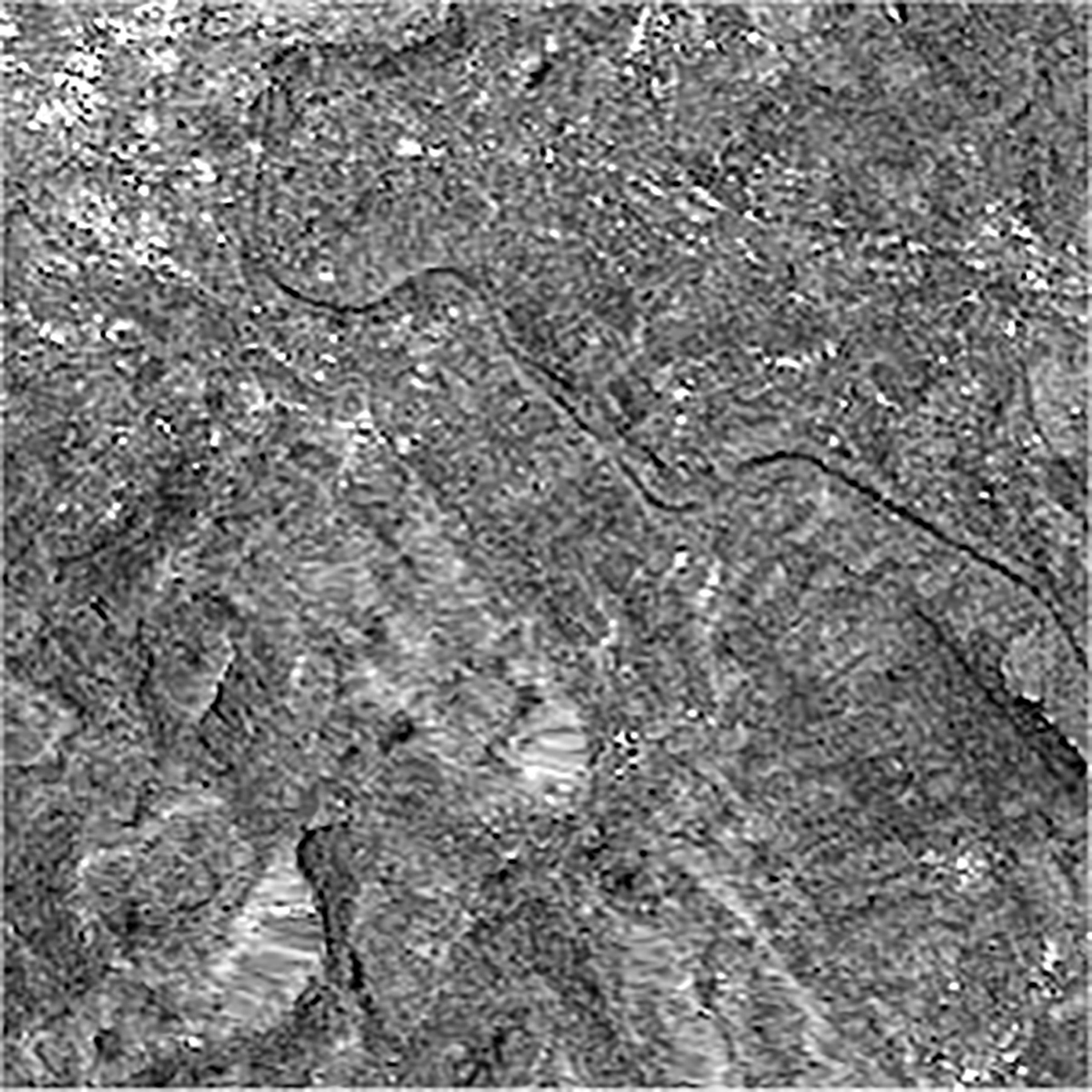}
	}
	\subfigure[]{
		\centering
		\includegraphics[width=0.3\linewidth]{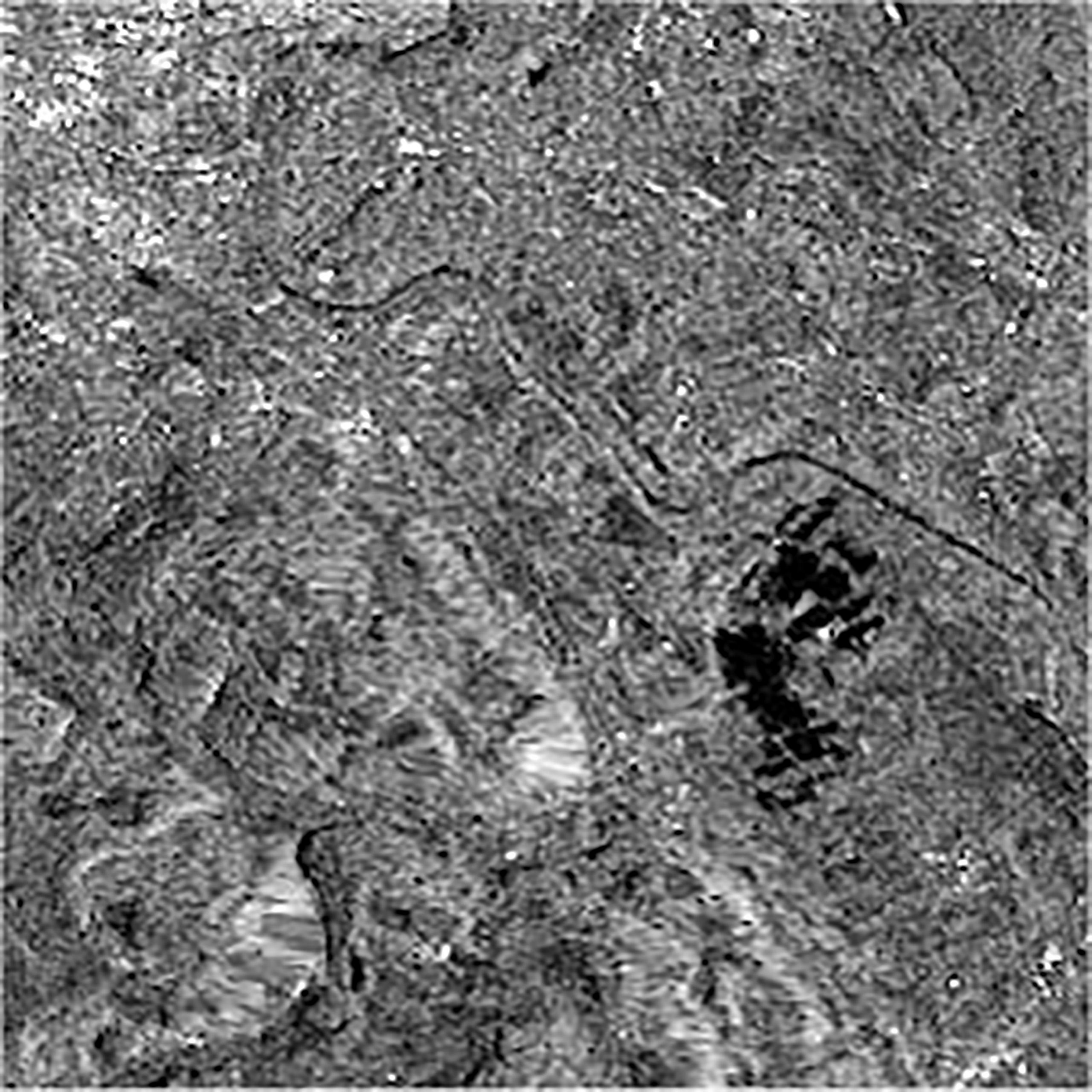}
	}
	\subfigure[]{
		\centering
		\includegraphics[width=0.3\linewidth]{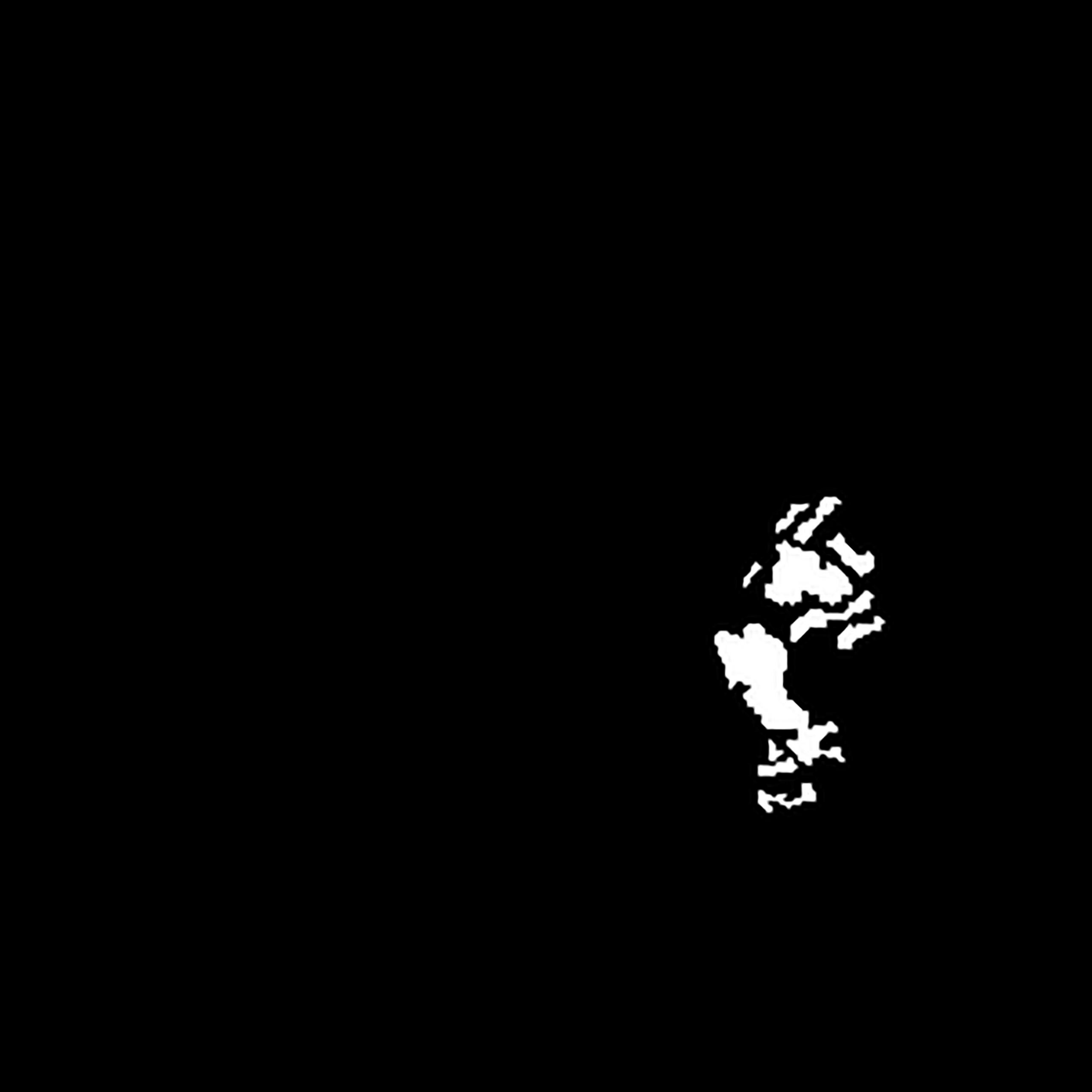}
	}
	\caption{ Images relating to the Bern: (a) Image acquired in September 1995. (b) Image acquired in July 1996. (c) Ground truth.}
	\label{figure4}
	
\end{figure}

 The second dataset (shown in Fig.\ref{figure4}(a) and (b)) consists of bi-temporal SAR images, acquired by the European Remote Sensing 2 satellite SAR sensor at the region of Bern, Switzerland, in April and May 1999,respectively. The ground truth is shown in Fig.\ref{figure4} (c). The size of two images is $301 \times 301$ pixels.

\begin{figure}[htbp]
	\centering
	
	\subfigure[]{
		\centering
		\includegraphics[width=0.3\linewidth]{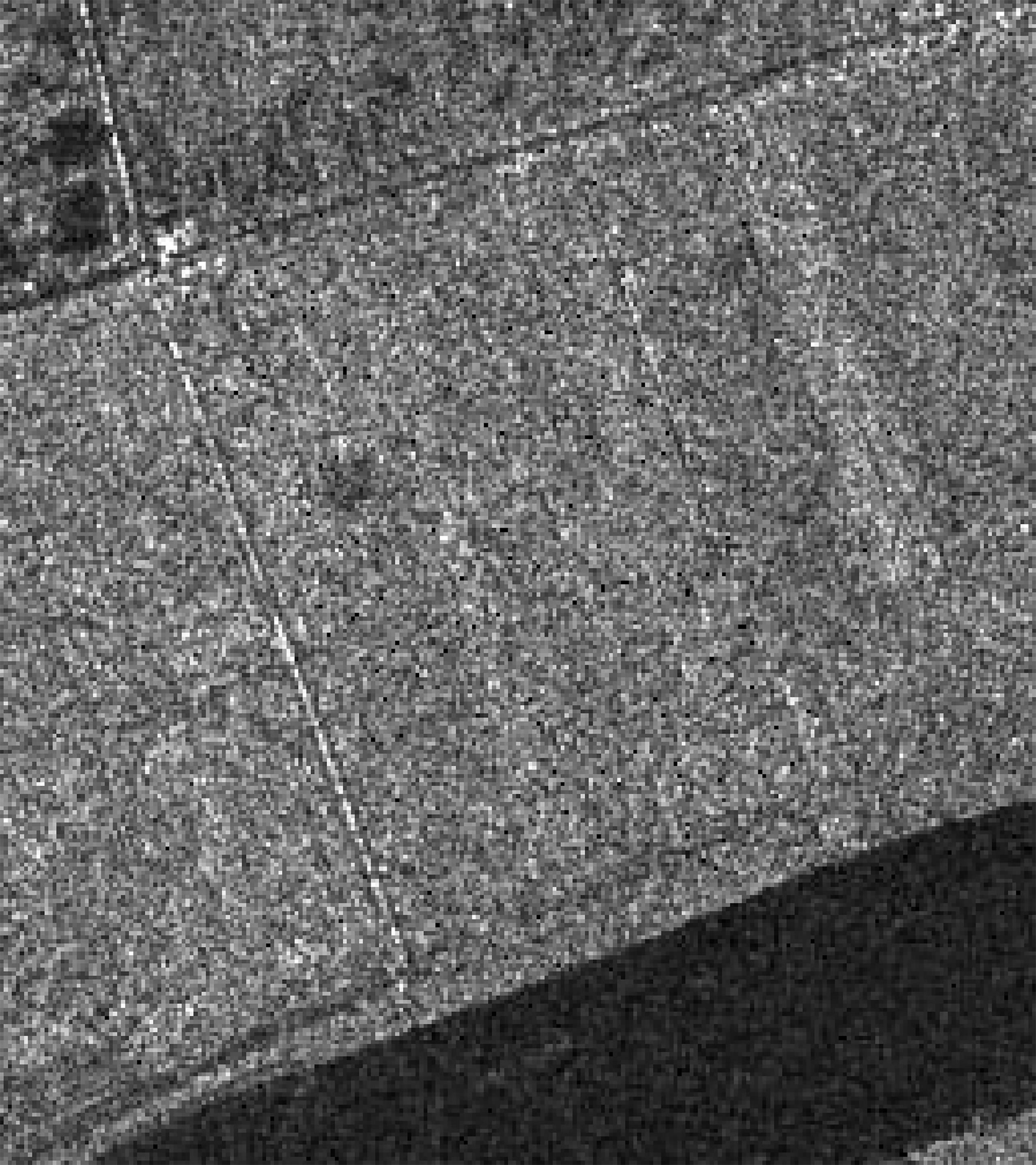}
	}
	\subfigure[]{
		\centering
		\includegraphics[width=0.3\linewidth]{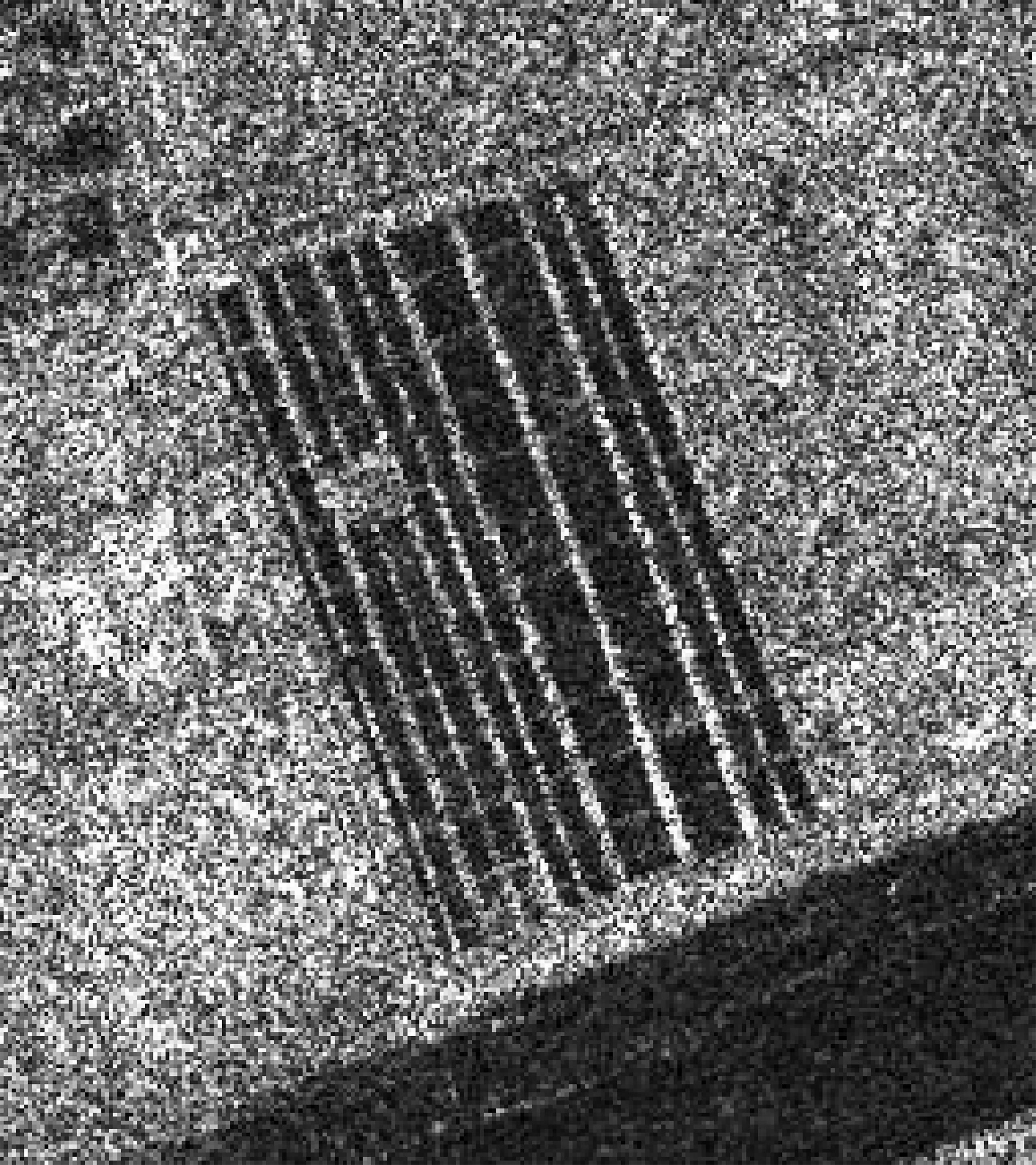}
	}
	\subfigure[]{
		\centering
		\includegraphics[width=0.3\linewidth]{fig_5_b.pdf}
	}
	\caption{ Images relating to the Yellow River: (a) Image acquired in September 1995. (b) Image acquired in July 1996. (c) Ground truth.}
	\label{figure5}
	
\end{figure}

The third dataset (shown in Fig.\ref{figure5}(a) and (b))  consists of bi-temporal SAR images, acquired by Radarsat-2 at the region of Yellow River Estuary in China in June 2008 and June 2009, respectively. The ground truth is shown in Fig.\ref{figure5}(c). The size of two images is $257 \times 289$ pixels.

\par
\begin{figure}[htbp]
	\centering
	
	\subfigure[]{
		\centering
		\includegraphics[width=0.3\linewidth]{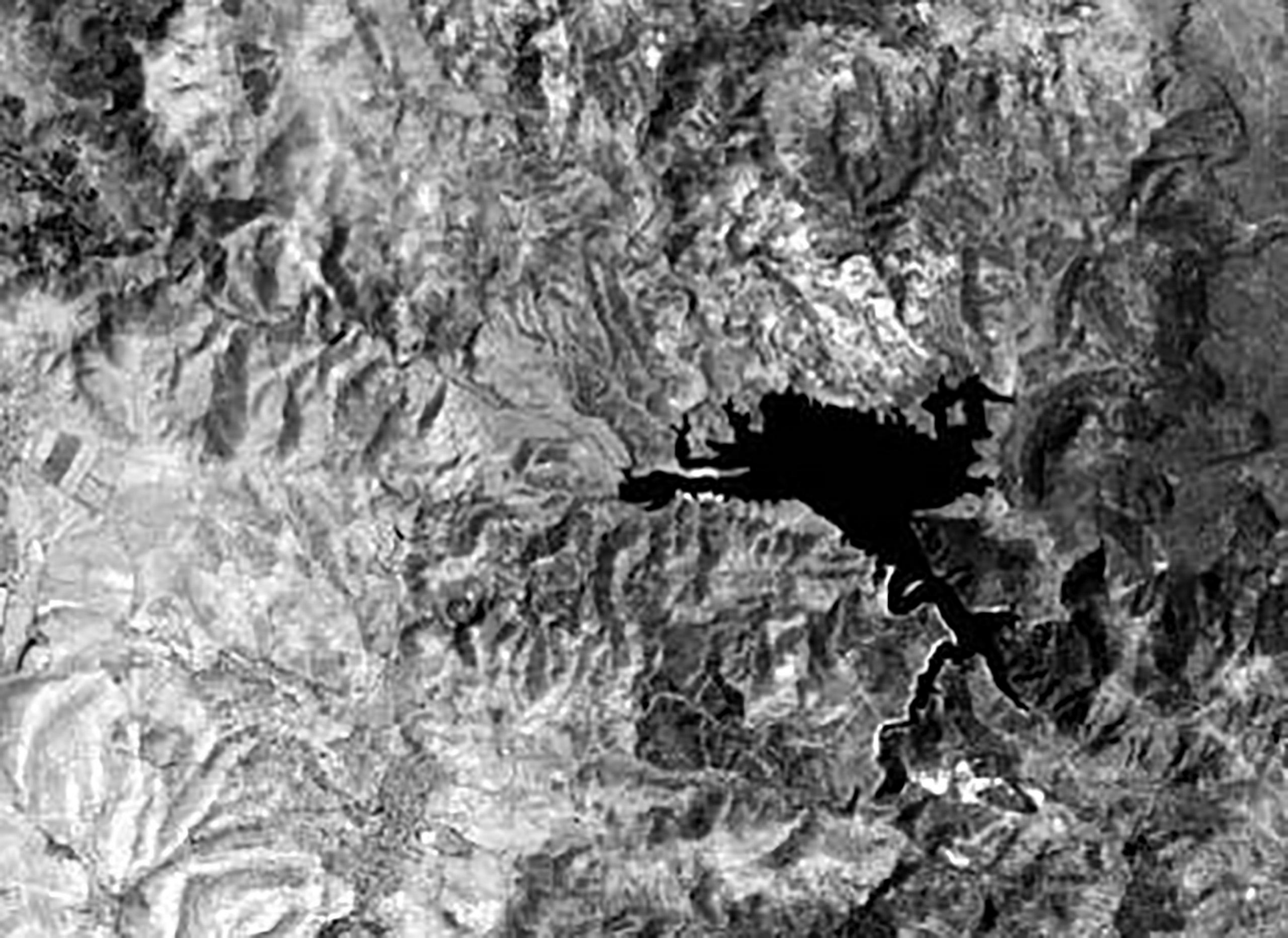}
	}
	\subfigure[]{
		\centering
		\includegraphics[width=0.3\linewidth]{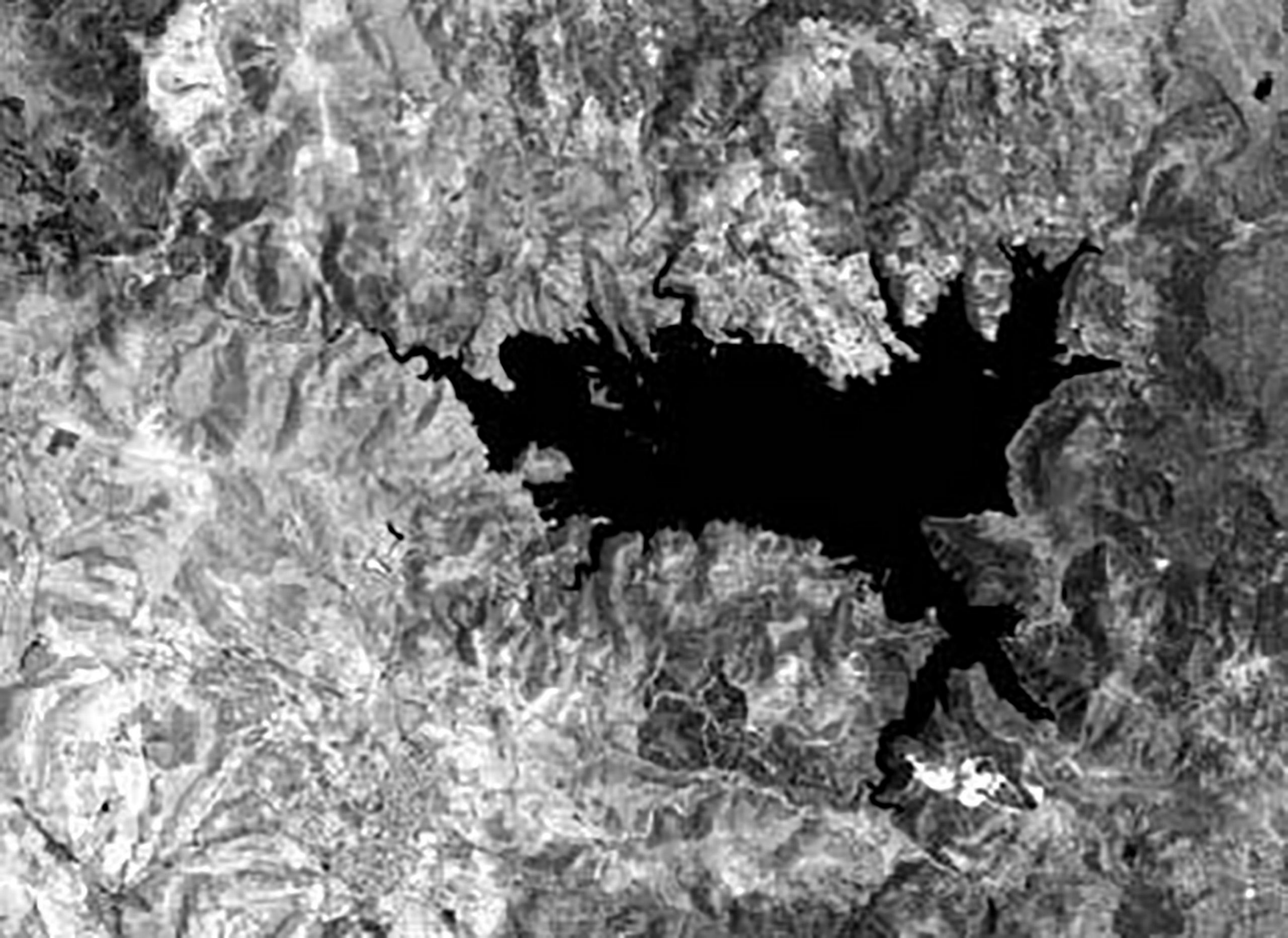}
	}
	\subfigure[]{
		\centering
		\includegraphics[width=0.3\linewidth]{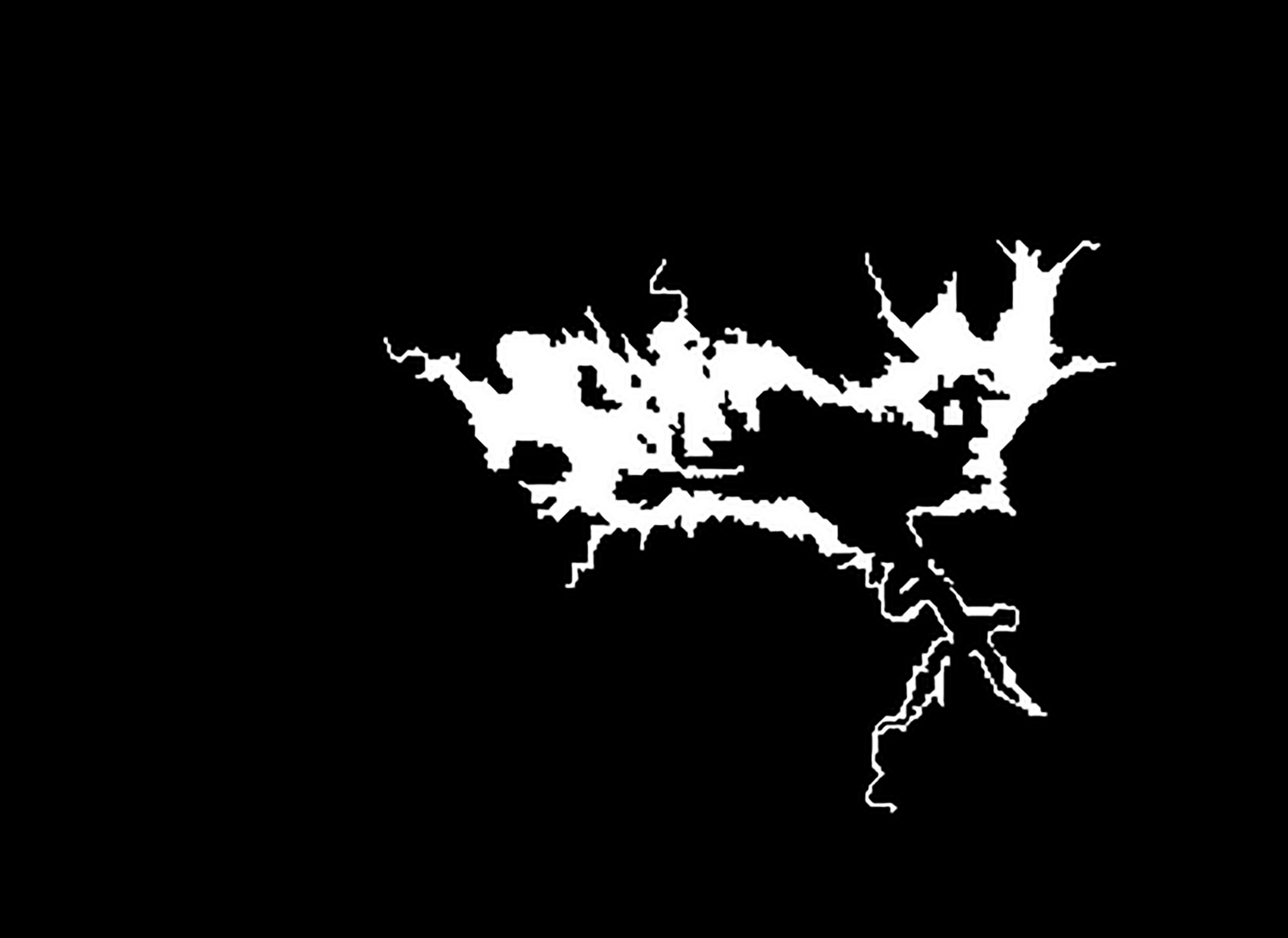}
	}
	\caption{ Images relating to the Sardinia: (a) Image acquired in September 1995. (b) Image acquired in July 1996. (c) Ground truth.}
	\label{figure6}
	
\end{figure}

 The forth dataset (shown in Fig.\ref{figure6} (a) and (b)) consists of bi-temporal TM images, acquired by Landsat-5 at the area of Italy in September 1995 and July 1996, respectively. The ground truth is shown in Fig.\ref{figure6} (c). The size of two TM images are $412 \times 300$ pixels.

\par For the quantitative evaluation of the results of the USCNN, false negative ($FN$, the number of changed pixels wrongly classified into unchanged class), false positive ($FP$, the number of unchanged pixels wrongly classified into changed class), overall errors ($OE$, the sum of $FP$ and $FN$), $PCC$ and $Kappa$ are used as criteria \cite{25} to illustrate the performance of the experiments on four real datasets. To be specific,$PCC$ is defined as:

\begin{equation}\label{13}
PCC=(TP+TN)/N
\end{equation}

\noindent where $N$ is the number of all pixels in the image. $TP$ is the number of changed pixels correctly classified into changed classes. $TN$ is the number of unchanged pixels correctly classified into unchanged classes.

$Kappa$ is defined as:

\begin{equation}\label{14}
Kappa=(PCC-PRE)/(1-PRE)
\end{equation}

\noindent where

\begin{equation}\label{15}
PRE=\frac{(TP+FP)Mc+(FN+TN)Mu}{N^2}
\end{equation}

\noindent $Mu$ and $Mc$ denote the actual number of pixels in changed class and unchanged class, respectively.

\subsection{Analysis of parameter k}\label{subsect3.2}

\begin{figure}[t]
	\centering
	\includegraphics[width=10cm,height=7.5cm]{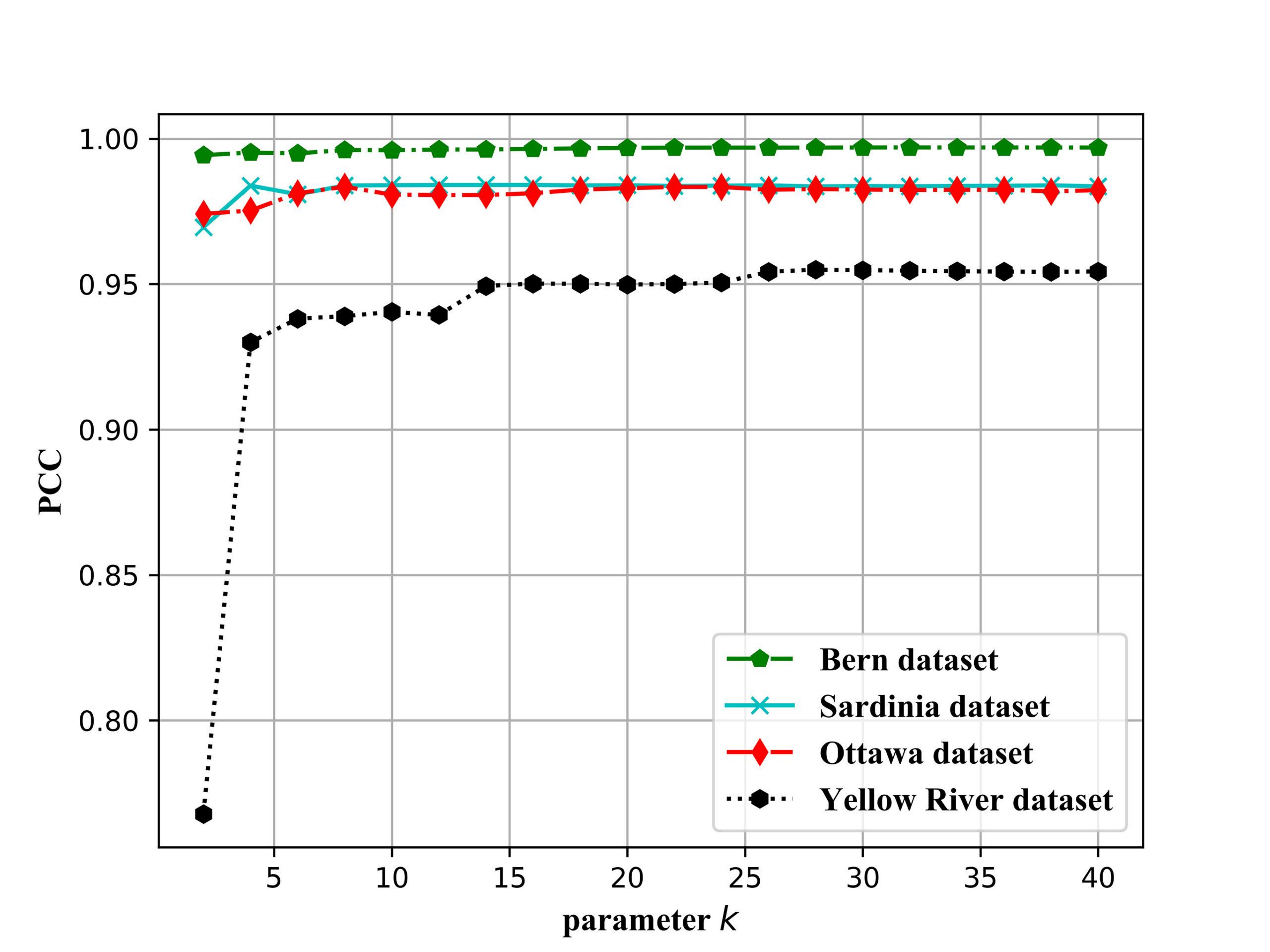}
	\caption{Relationship between $k$ and $PCC$ on four data sets.}
	\label{figure8}
\end{figure}

Before giving the final results of USCNN, the parameter $k$ is discussed in this subsection. In order to find the optimal parameter $k$, we set $k$ to 2,4,6,...,40 to indicate the relationship between $k$ and $PCC$. To suppress noise, we perform a logarithmic operation on the original images during preprocessing. We train our model with RMSprop optimizer with a base learning rate of 0.01 and set the maximum number of epochs to 100. The relationship between $k$ and $PCC$ is shown in Fig.\ref{figure8}.  As we can see from Fig.\ref{figure8}, when $k$ is set to be greater than or equal to 2, $PCC$ increases slowly as $k$ increases on the Bern, Sardinia, and Ottawa datasets. For the Yellow River dataset, when $k$ increases from 2 to 16, $PCC$ increases significantly, but after that, $PCC$ increases quite slowly until $k$ reaches about 30. While $k$ exceeds 30, $PCC$ shows a rather slow downward trend. Therefore, we can observe that $k = 30$ is a good choice for each dataset.

\subsection{Result on the different datasets}\label{subsect3.3}
In this subsection, we present the final experiment results on four different real datasets. According to the parameters analysis in Part B, we set $k=30$. Then, we quantify the impact of the proposed method on change detection results via conducting four groups of experiments compared with four other classic algorithms, including LMR, principal component analysis (PCA) \cite{26}, PCANet \cite{8}, extreme learning machine (ELM) \cite{9} and CDML\cite{13}. The above mentioned algorithms are all implemented with default parameters provided in \cite{26,8,9}. In addition, k-Means clustering algorithm is applied to segment the difference maps generated by LMR, PCA and CDML and USCNN. PCANet and ELM do not need to use any clustering algorithms because there is no intermediate step to generate a difference map. It is worth noting that the LMR, PCA and USCNN are three completely unsupervised methods, which do not need any training samples, while PCANet, ELM  and CDML often need to use some kind of pre-detection mechanism to select the appropriate training samples.

\subsubsection{Result on the Ottawa dataset}\label{subsunsect3.3.1}

\begin{figure}[htbp]
	\centering
	
	\subfigure[]{
		\centering
		\includegraphics[width=0.3\linewidth]{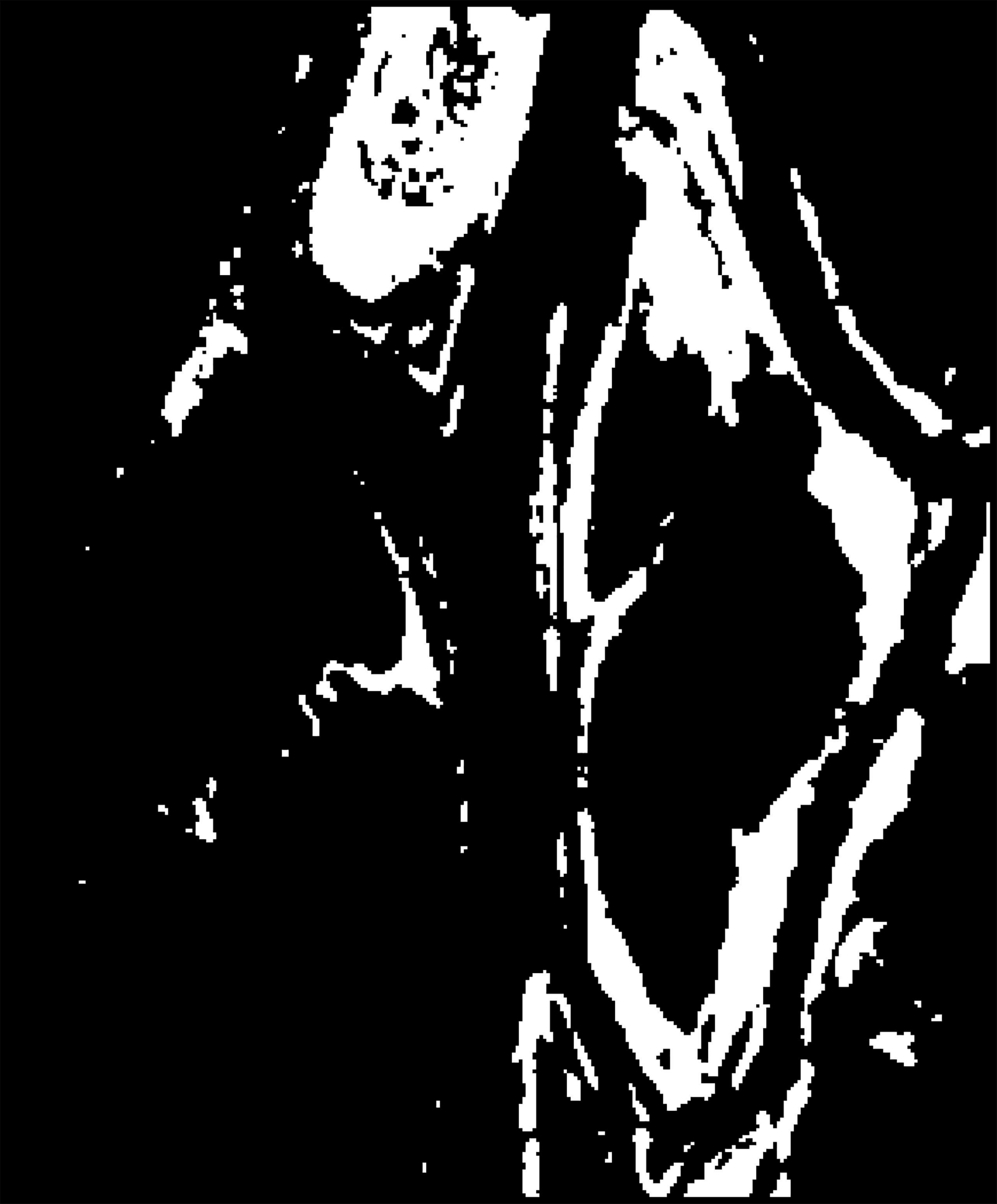}
	}
	\subfigure[]{
		\centering
		\includegraphics[width=0.3\linewidth]{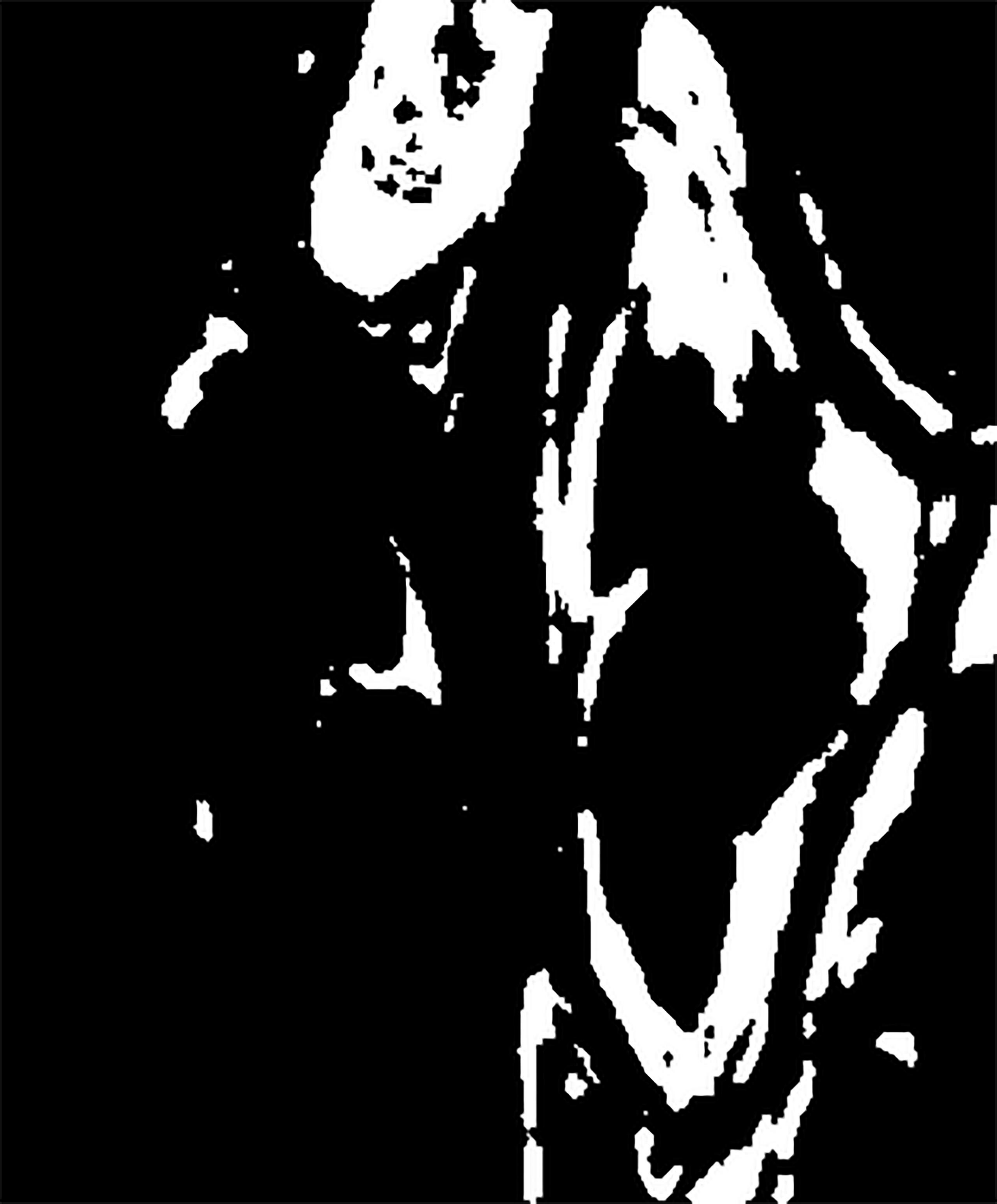}
	}
	\subfigure[]{
		\centering
		\includegraphics[width=0.3\linewidth]{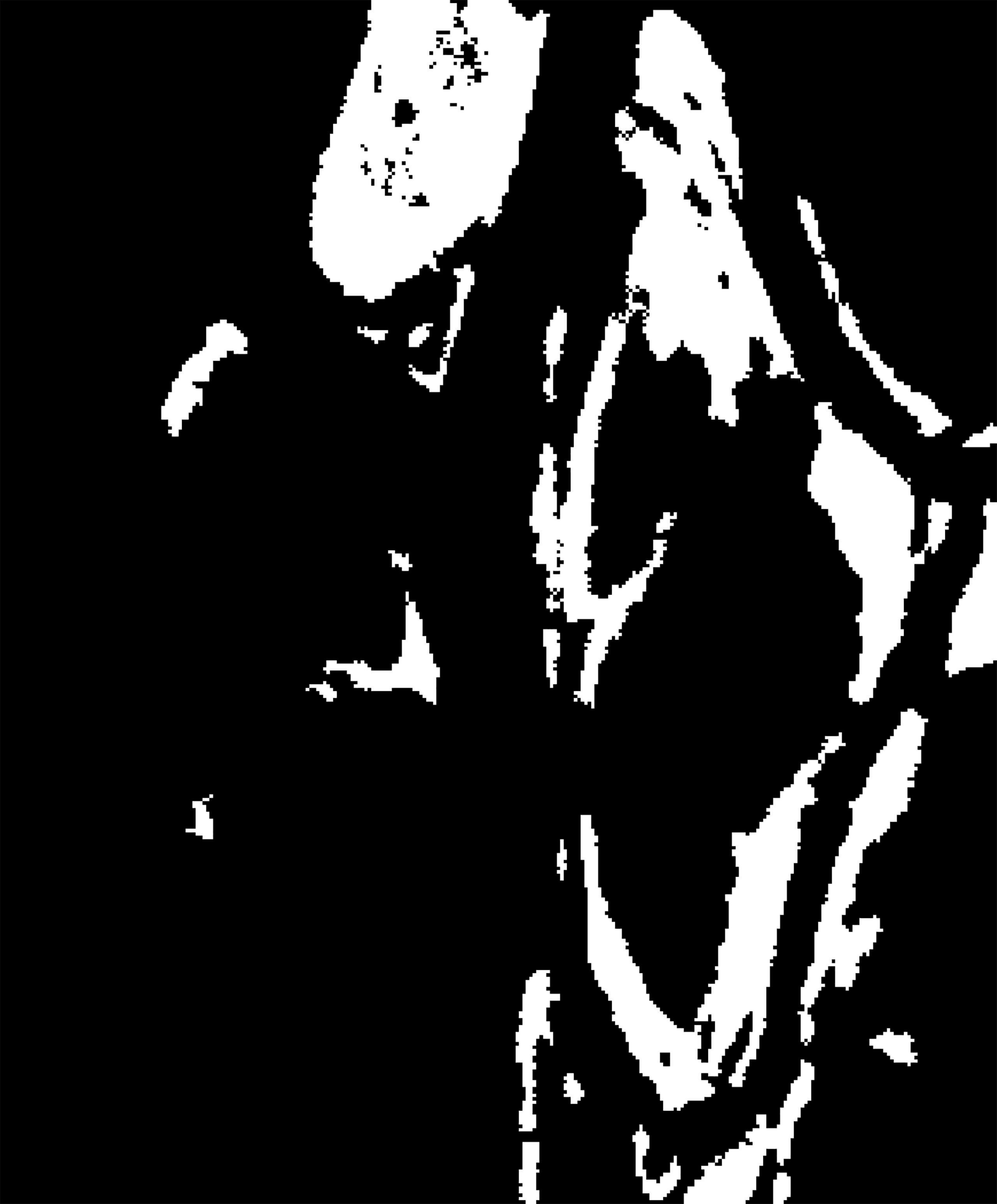}
	}
	\vfill
	\subfigure[]{
		\centering
		\includegraphics[width=0.3\linewidth]{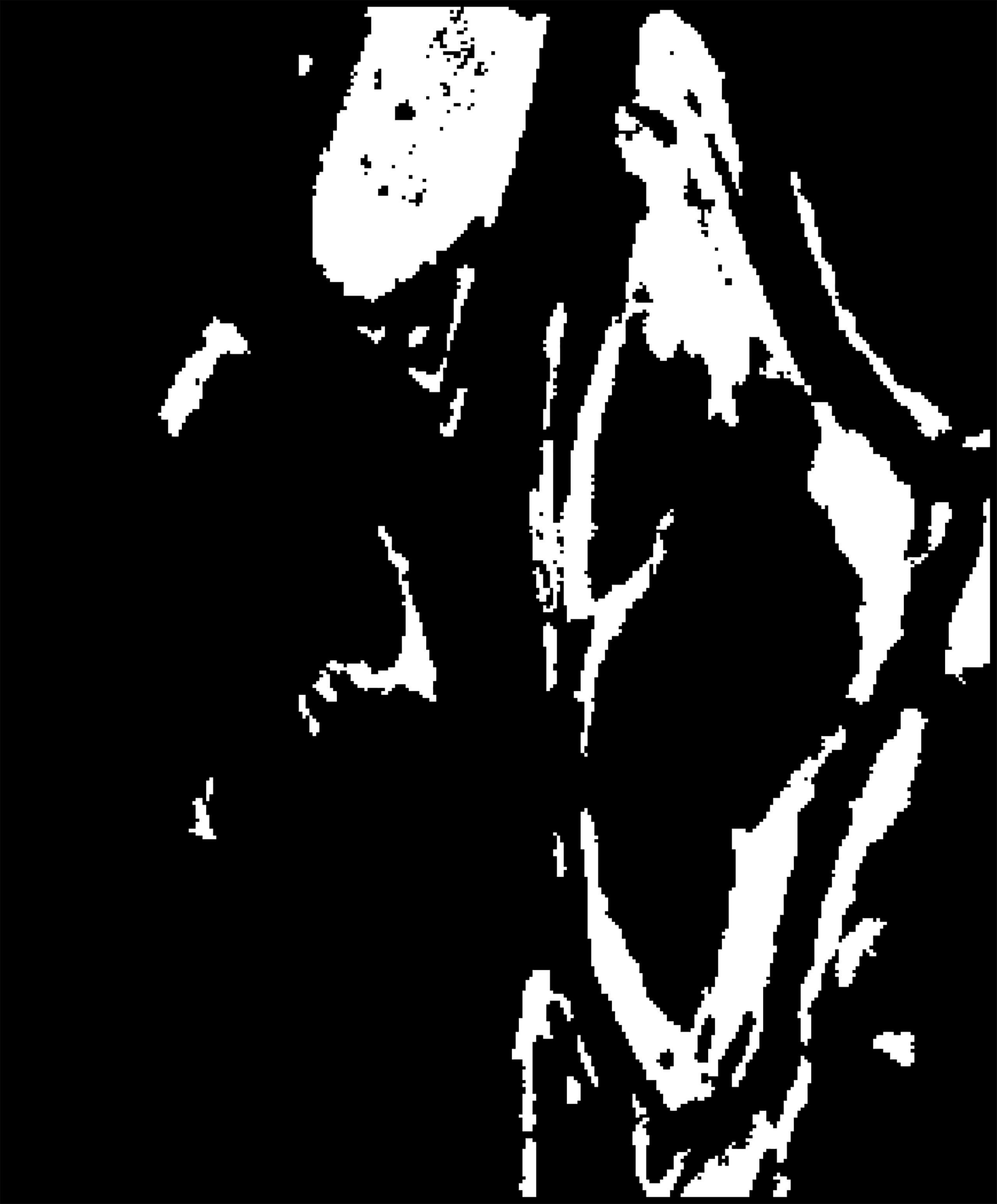}
	}
	\subfigure[]{
		\centering
		\includegraphics[width=0.3\linewidth]{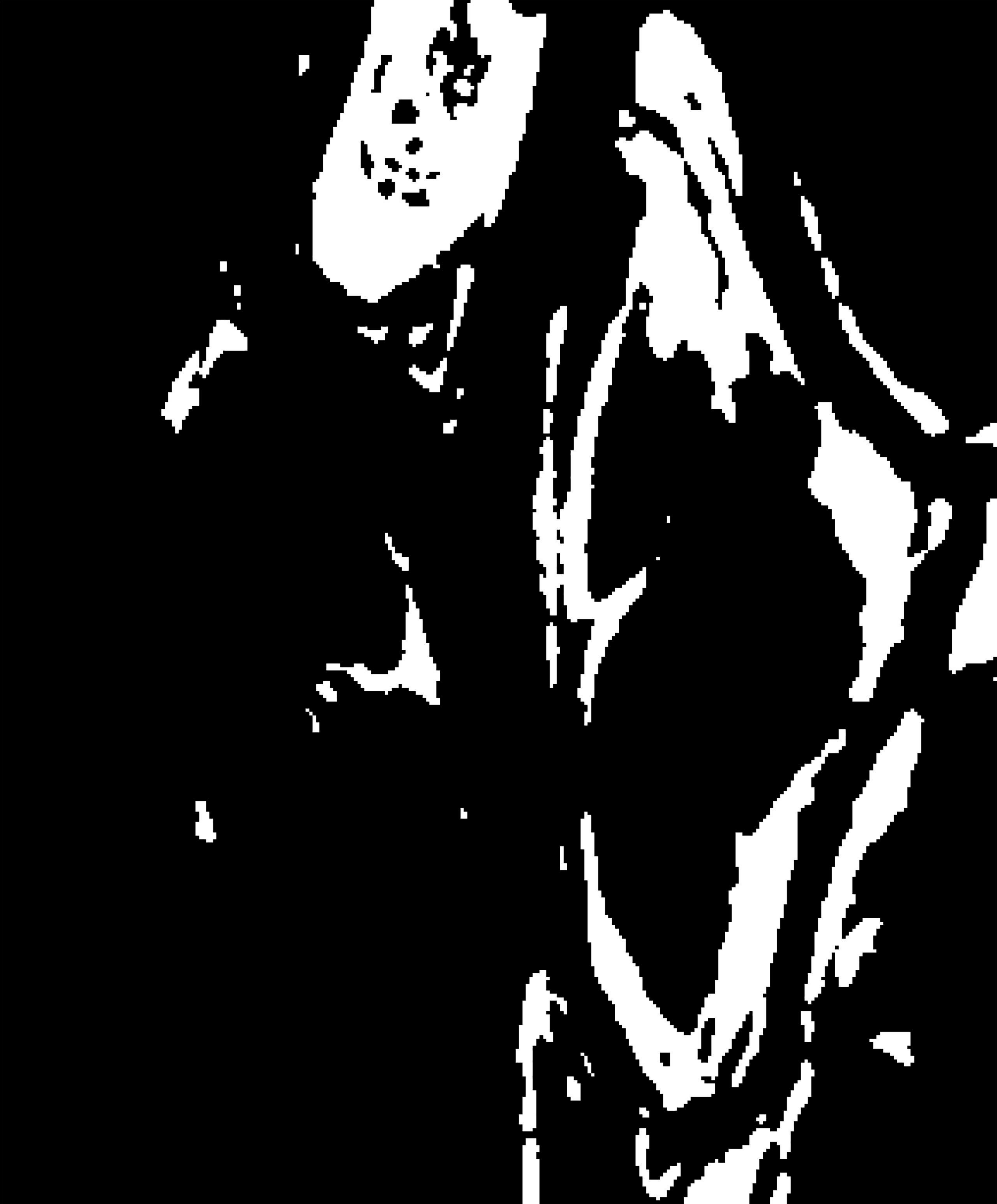}
	}
	\subfigure[]{
		\centering
		\includegraphics[width=0.3\linewidth]{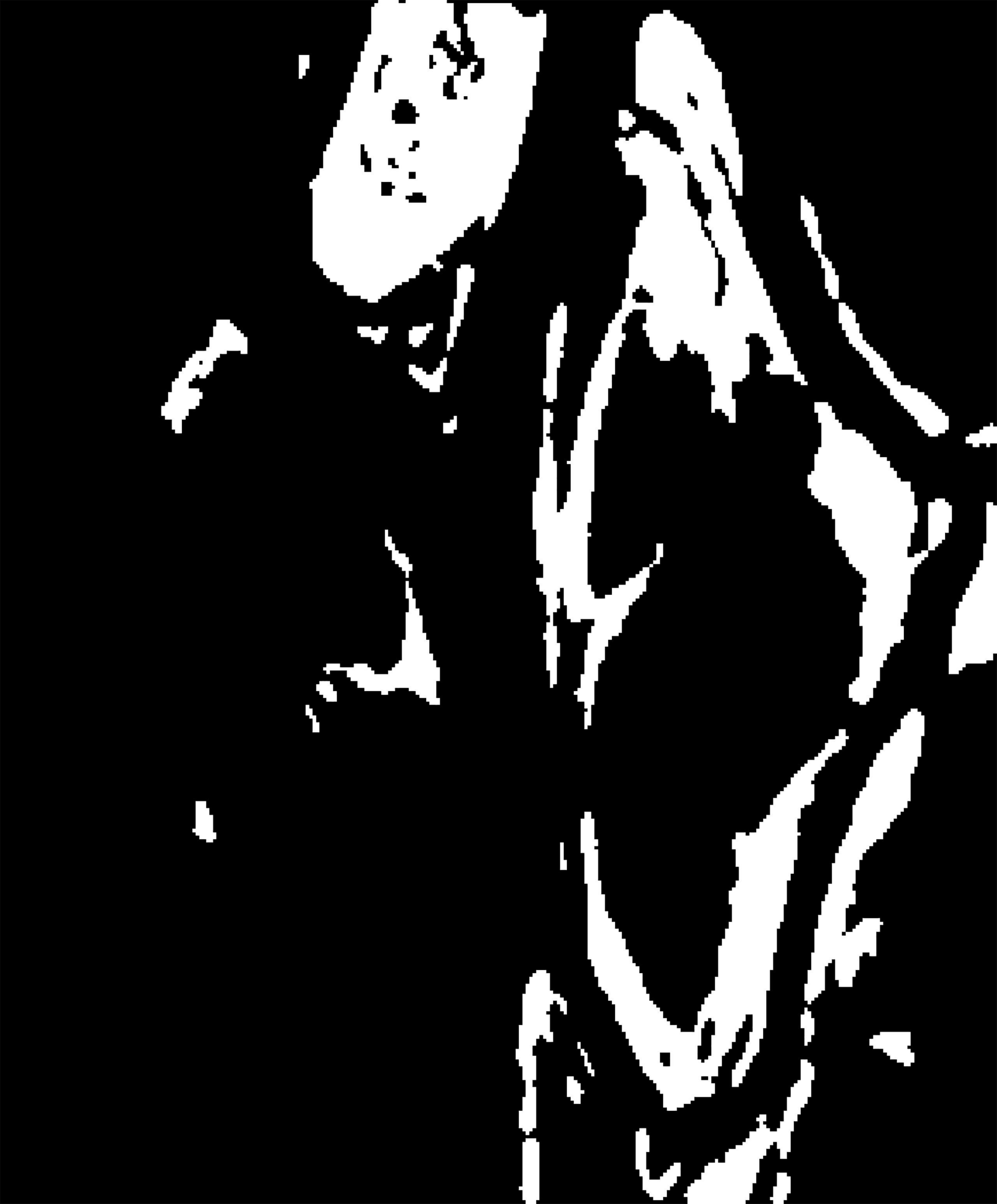}
	}
	
	\caption{Binary maps for Ottawa dataset obtained by different methods. (a) Binary map by LMR. (b) Binary map by PCA. (c) Binary map by PCANet. (d) Binary map by ELM. (e) Binary map by CDML. (f) Binary map by USCNN.}
	\label{figure12}
	
\end{figure}

Fig.\ref{figure12} and Table \ref{table4} present the results of Ottawa data set by LMR, PCA, PCANet, ELM ,CDML and USCNN. As shown in Fig.\ref{figure12}(a), the difference map by LMR looks noisy. In the results generated by PCA and CDML, there are a large amount of unchanged pixels that are classified as changed pixels. In contrast, the result maps generated by PCANet, ELM and USCNN are approximate to the ground truth. As shown in Table \ref{table4}, we can observe that USCNN yields the best $PCC$ value and $Kappa$ value on the Ottawa dataset, which shows the effectiveness of the proposed method.

\begin{table}[htbp]
	\centering
	\caption{Change detection indicators of different methods on the Ottawa dataset}
	\label{table4}
	\begin{tabular}{cccccc}
		\hline
		Methods & $FP$ & $FN$ & $OE$ & $PCC$ & $Kappa$\\
		\hline
		LMR    &719  &1522 &2241&	0.9779&	0.9153\\
		PCA	   &972  &1541 &2513&	0.9752&	0.9056\\
		PCANet &778	 &1077 &1855&	0.9817&	0.9308\\
		ELM    &565	 &1185 &1750&	0.9828&	0.9342\\
		CDML   &289  &1544 &1833&   0.9819& 0.9299\\
		USCNN  &577	 &1081 &1658&	\textbf{0.9837} &\textbf{0.9379}\\
		\hline
	\end{tabular}
\end{table}

\subsubsection{Result on the Bern dataset}\label{subsubsect3.3.2}

\begin{figure}[htbp]
	\centering
	
	\subfigure[]{
		\centering
		\includegraphics[width=0.3\linewidth]{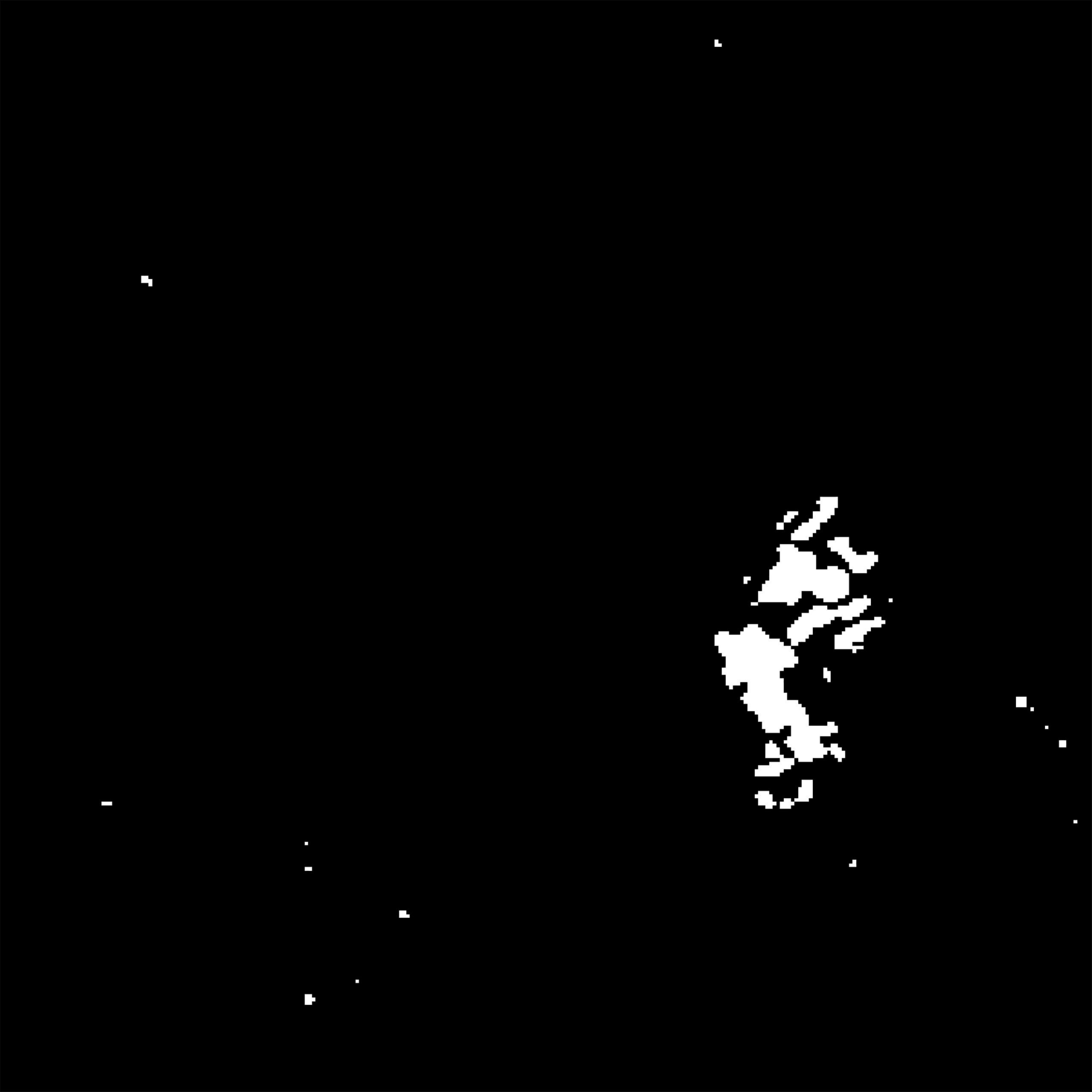}
	}
	\subfigure[]{
		\centering
		\includegraphics[width=0.3\linewidth]{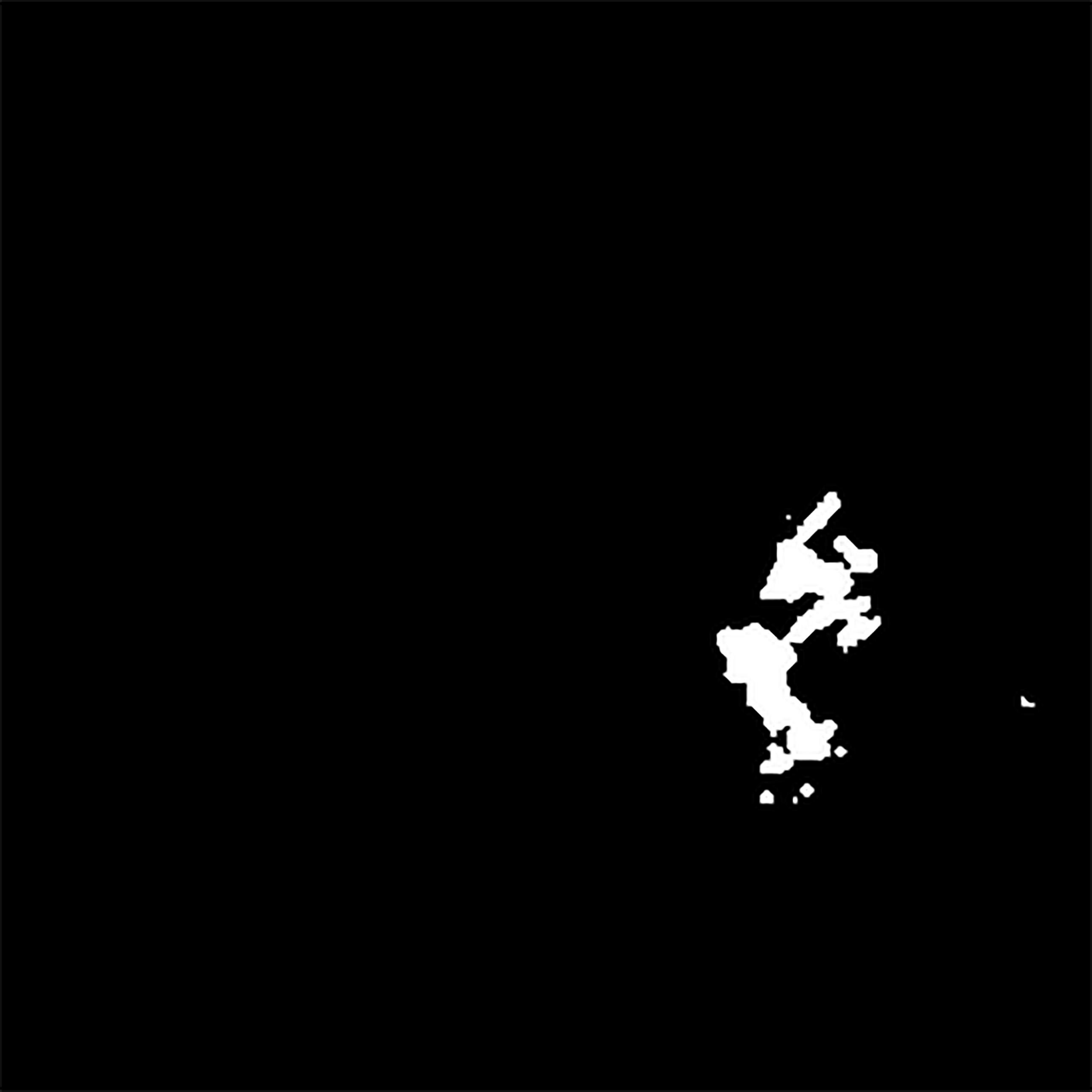}
	}
	\subfigure[]{
		\centering
		\includegraphics[width=0.3\linewidth]{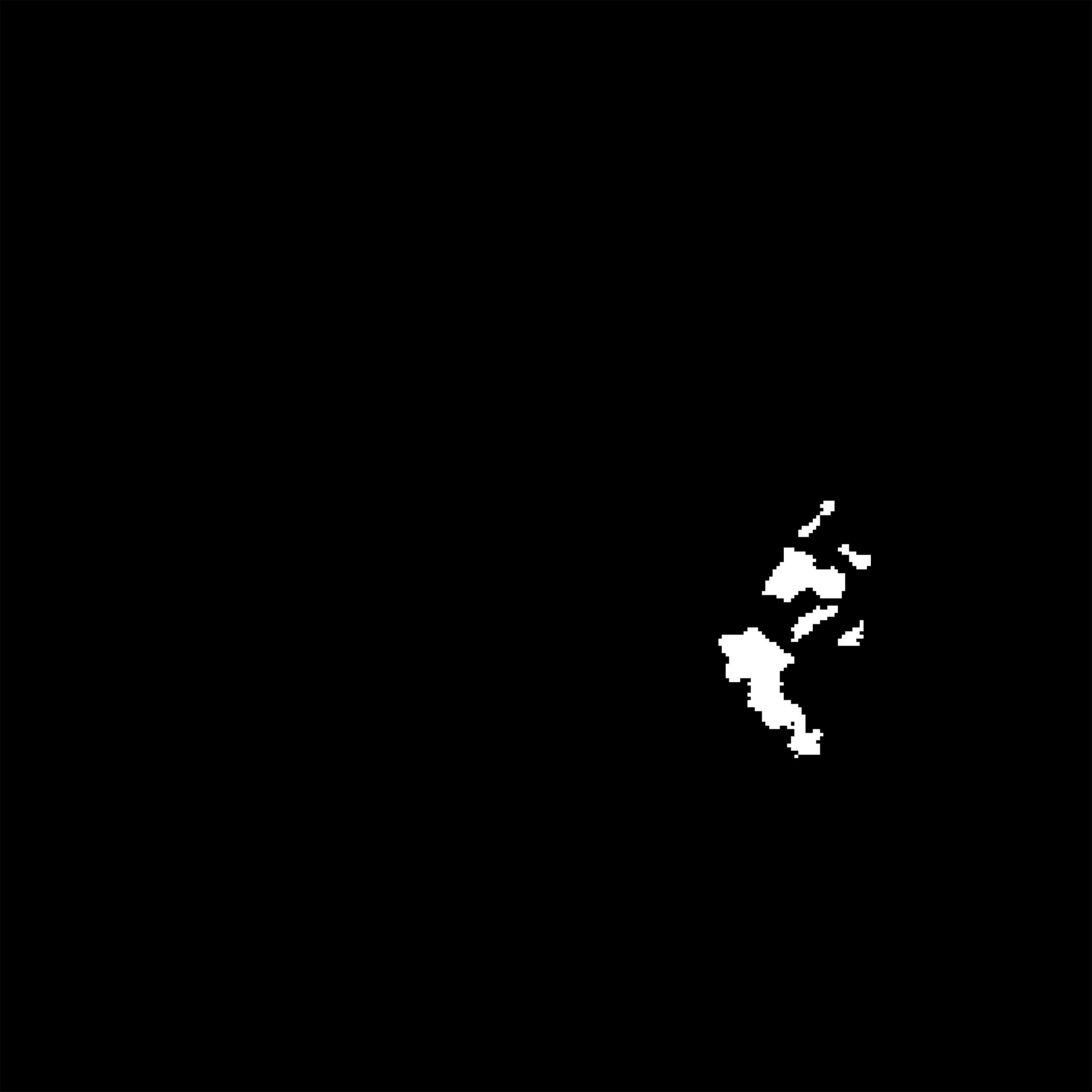}
	}
	\vfill
	\subfigure[]{
		\centering
		\includegraphics[width=0.3\linewidth]{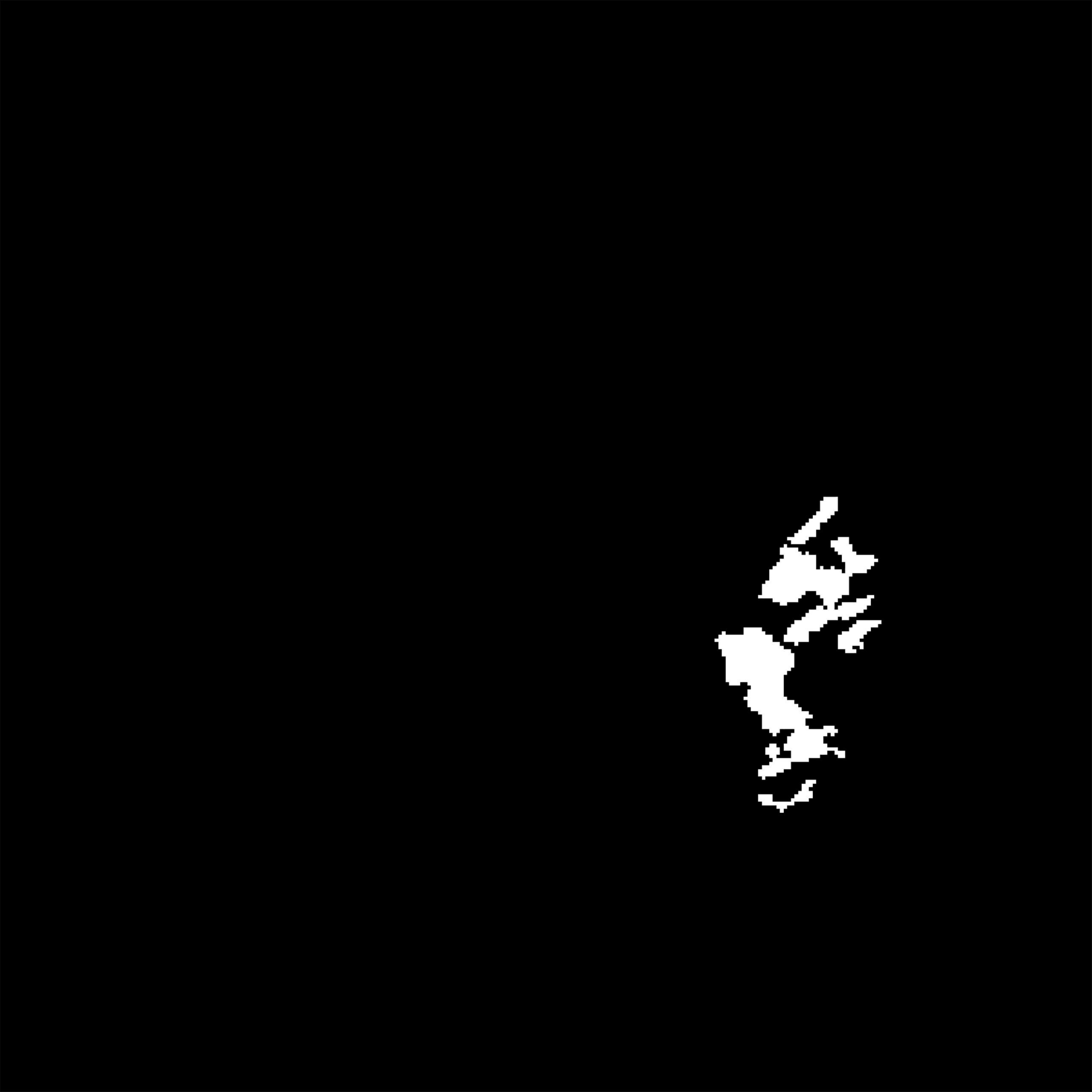}
	}
	\subfigure[]{
		\centering
		\includegraphics[width=0.3\linewidth]{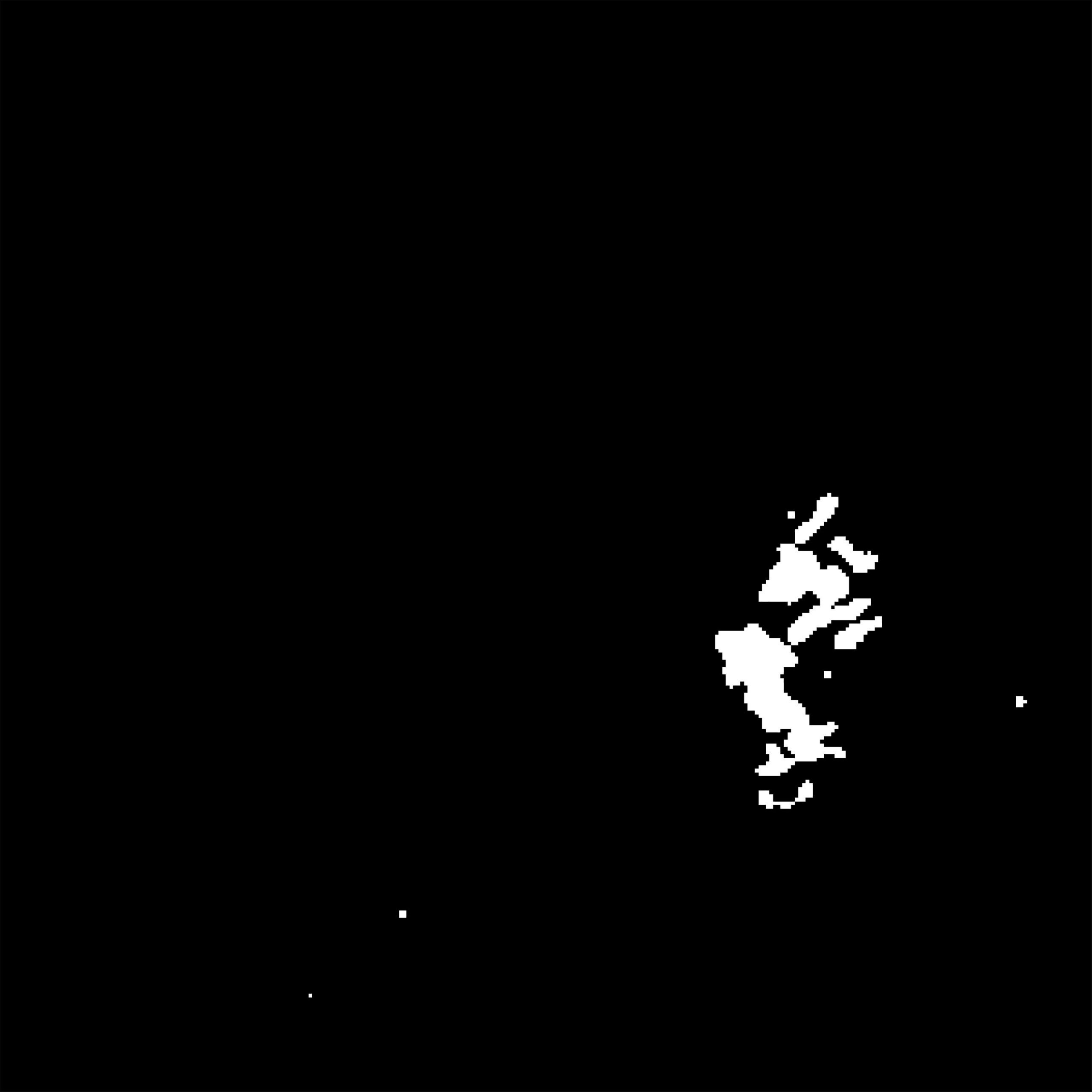}
	}
	\subfigure[]{
		\centering
		\includegraphics[width=0.3\linewidth]{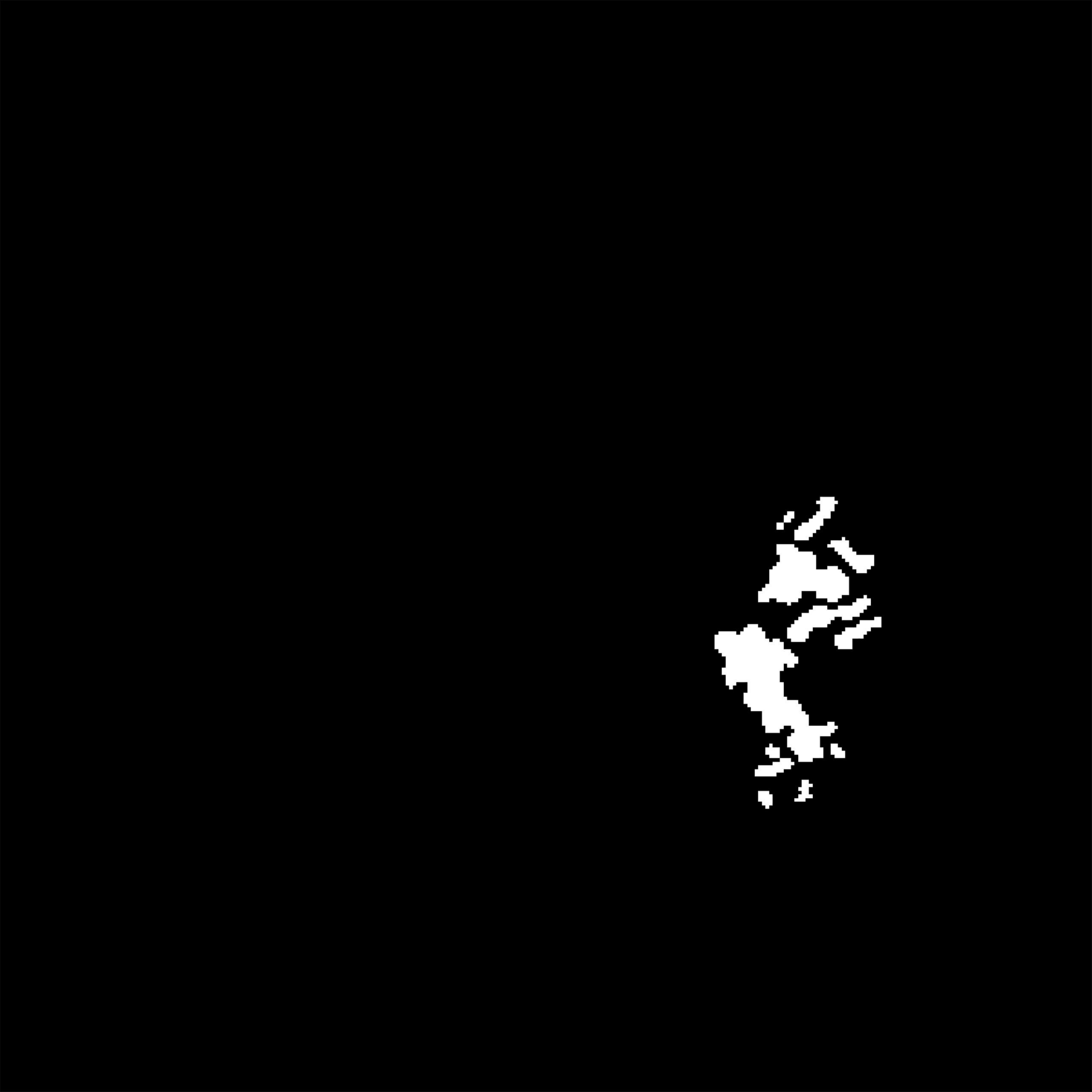}
	}
	
	\caption{Binary maps for Bern dataset obtained by different methods. (a) Binary map by LMR. (b) Binary map by PCA. (c) Binary map by PCANet. (d) Binary map by ELM. (e) Binary map by CDML. (f) Binary map by USCNN.}
	\label{figure9}
	
\end{figure}

Fig.\ref{figure9} presents the binary maps generated by LMR, PCA, PCANet, ELM, CDML and USCNN. The quantitative evaluation results are shown in Table \ref{table1}. In the result by LMR, there are a lot of isolated noises. Fig.\ref{figure9}(b), (d) and (e) effectively suppress the noise, but they lose too much detailed information. It can be observed that PCANet incorrectly classifies a large number of changed pixels into unchanged pixels, yielding the largest $FN$ and smallest $FP$ values. As shown in Fig.\ref{figure9}, compared with other methods, our method has better visual effects in terms of noise reduction and image detail preservation. Similarly, from the Table \ref{table1}, we can see that the $PCC$ value and $Kappa$ coefficient of USCNN are the largest among all, which indicates the proposed method performs better than other methods.

\begin{table}[htbp]
	\centering
	\caption{Change detection indicators of different methods on the Bern dataset}
	\label{table1}
	\begin{tabular}{cccccc}
		\hline
		Methods & $FP$ & $FN$ & $OE$ & $PCC$ & $Kappa$\\
		\hline
		LMR    &229 &110	&339    &0.9963	   &0.8585\\
		PCA	   &251	&123    &374	&0.9959    &0.8445\\
		PCANet &28  &446	&474    &0.9948	   &0.7470\\
		ELM    &153 &169    &322    &0.9964	   &0.8578\\
		CDML   &199 &106    &305    &0.9966    &0.8714\\
		USCNN & 118	&147	&265	&\textbf{0.9971}	&\textbf{0.8823}\\
		\hline
	\end{tabular}
\end{table}

\subsubsection{Result on the Yellow River dataset}\label{subsubsect3.3.3}

\begin{figure}[htbp]
	\centering
	
	\subfigure[]{
		\centering
		\includegraphics[width=0.3\linewidth]{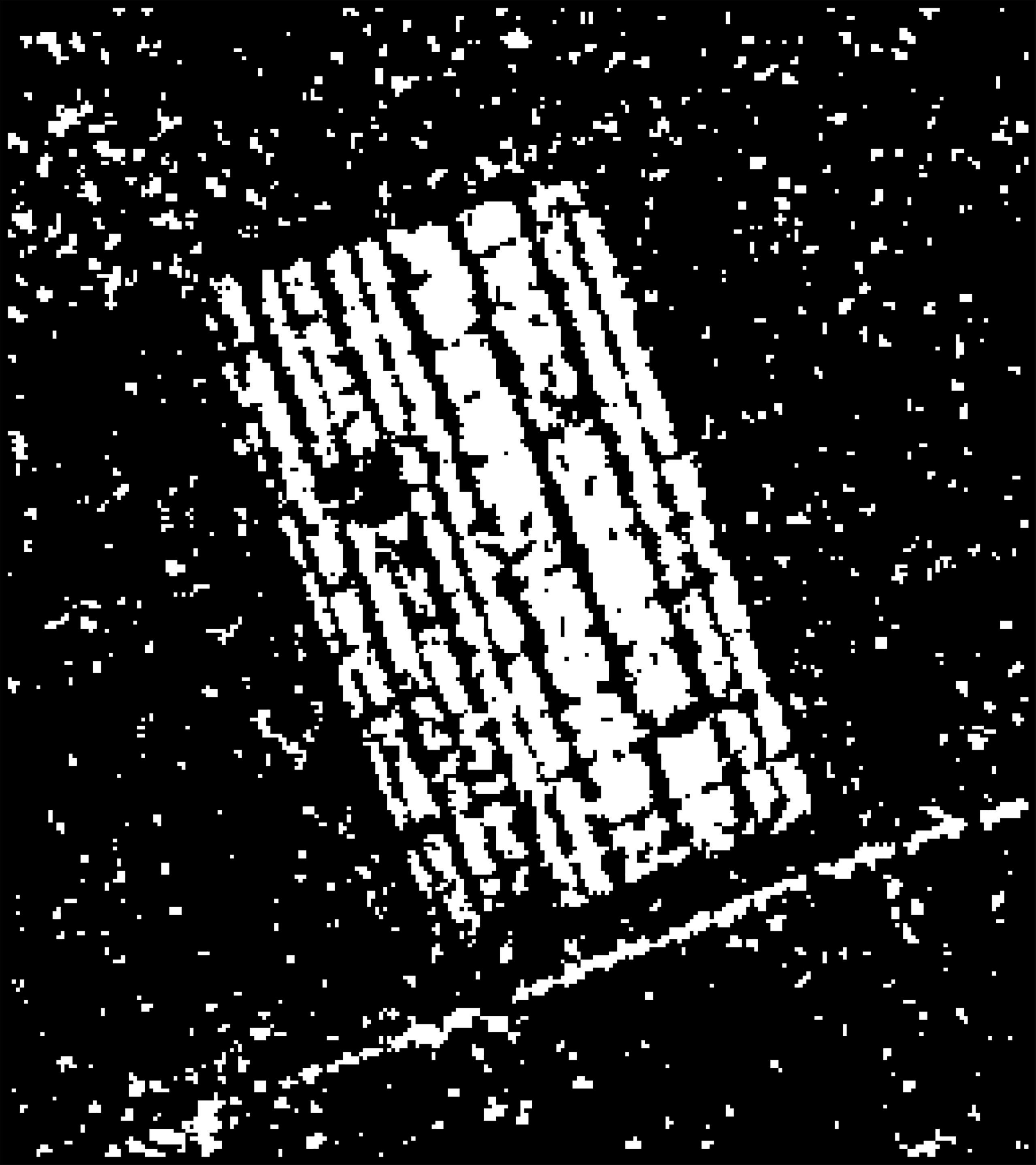}
	}
	\subfigure[]{
		\centering
		\includegraphics[width=0.3\linewidth]{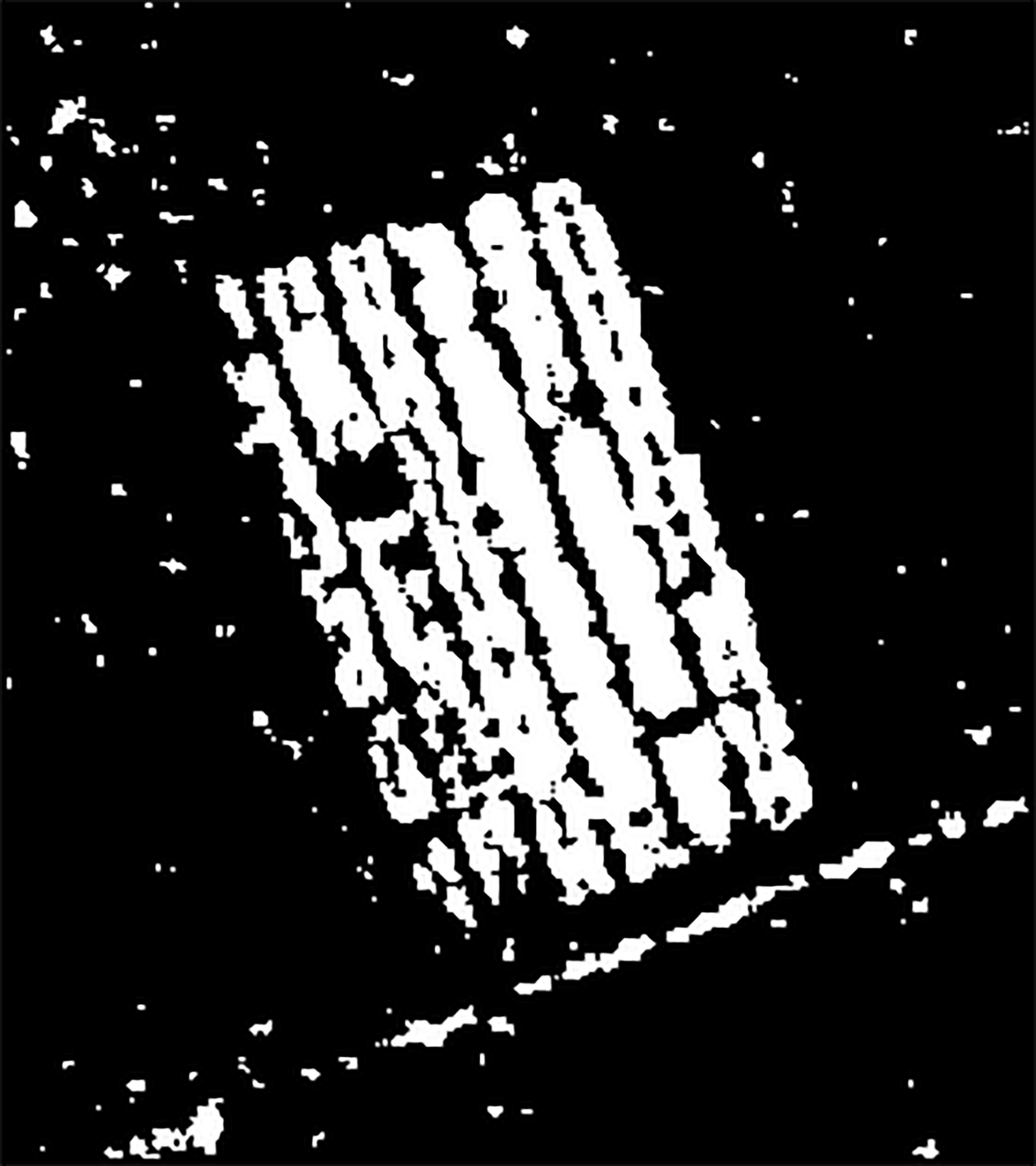}
	}
	\subfigure[]{
		\centering
		\includegraphics[width=0.3\linewidth]{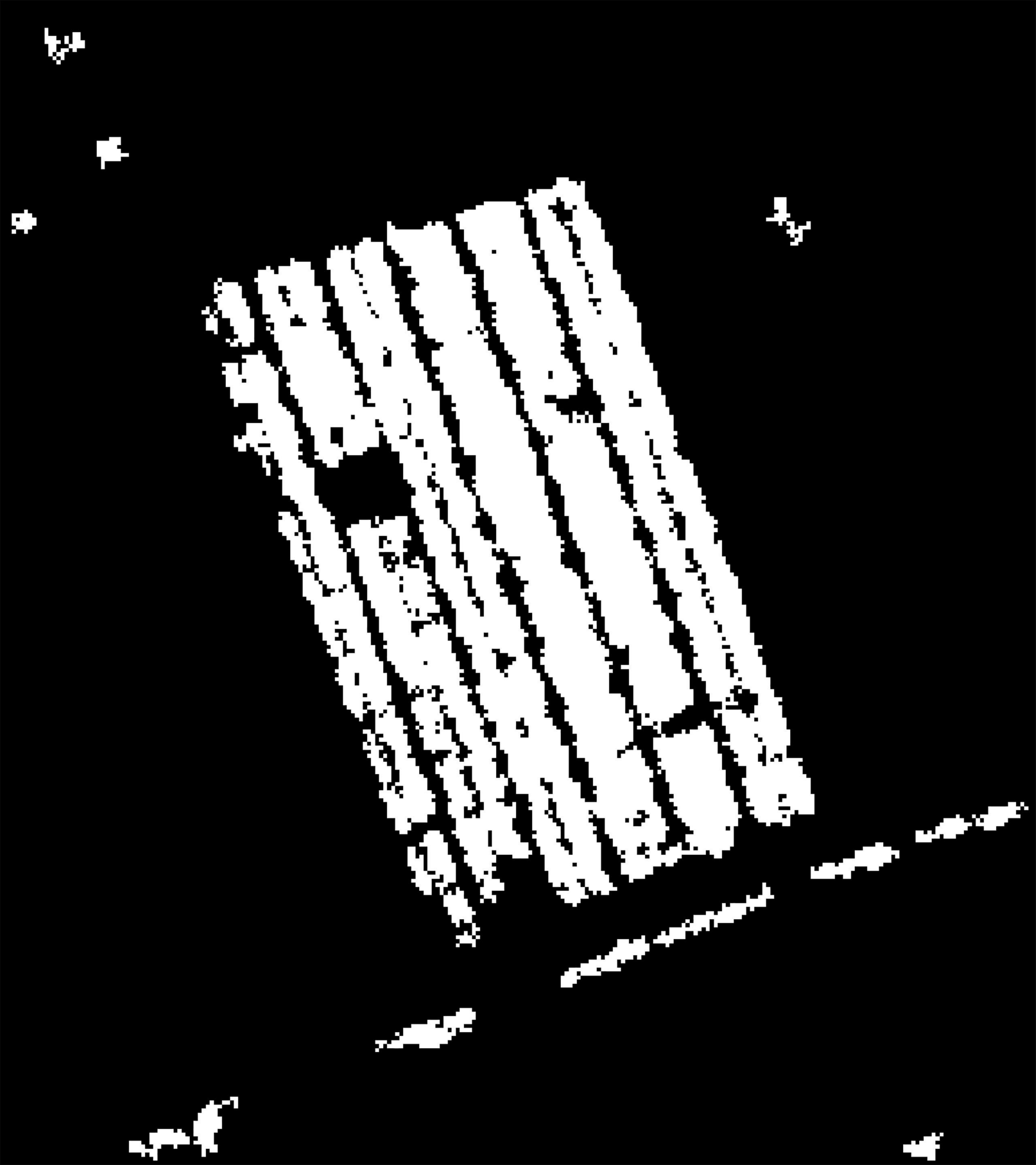}
	}
	\vfill
	\subfigure[]{
		\centering
		\includegraphics[width=0.3\linewidth]{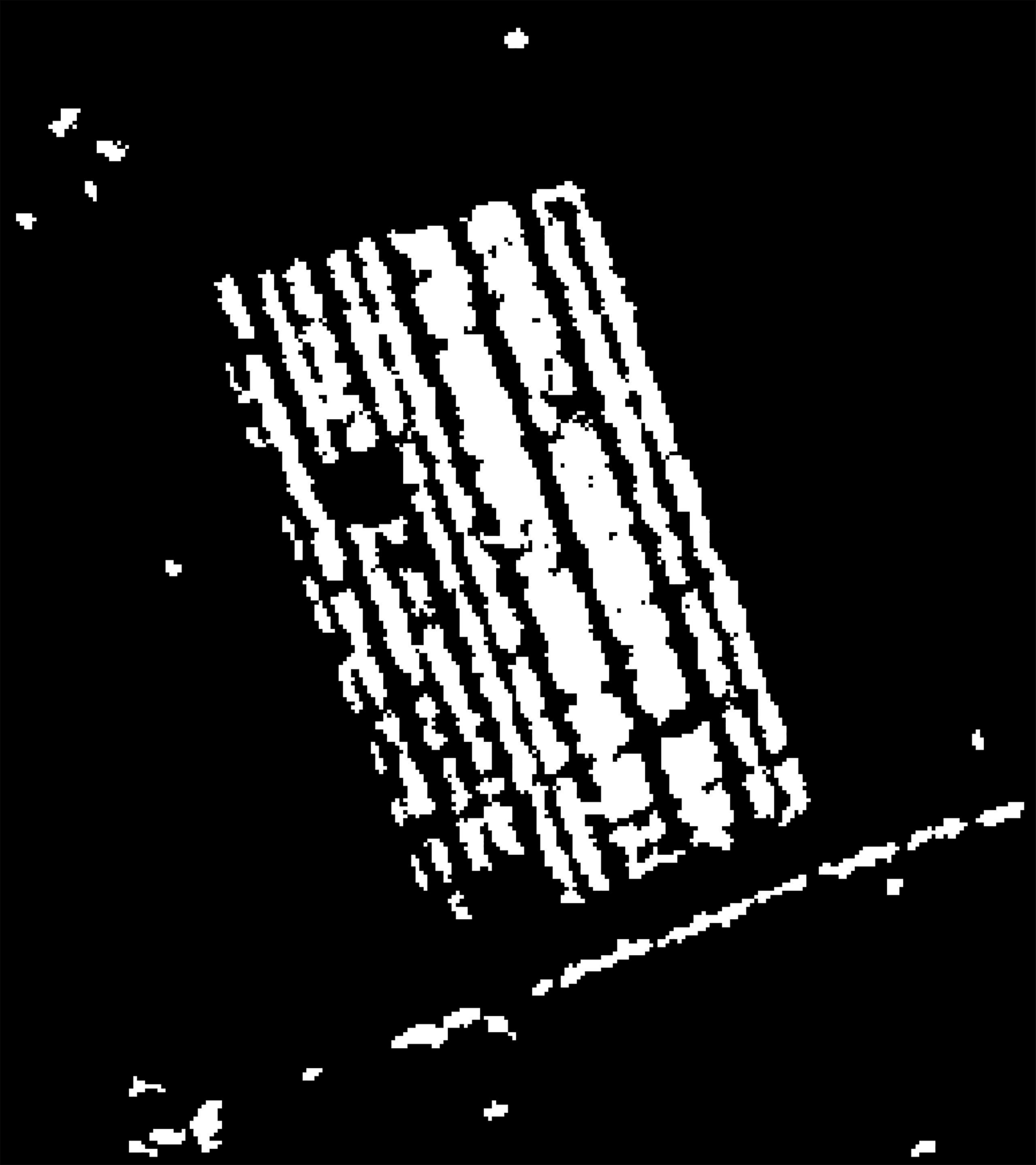}
	}
	\subfigure[]{
		\centering
		\includegraphics[width=0.3\linewidth]{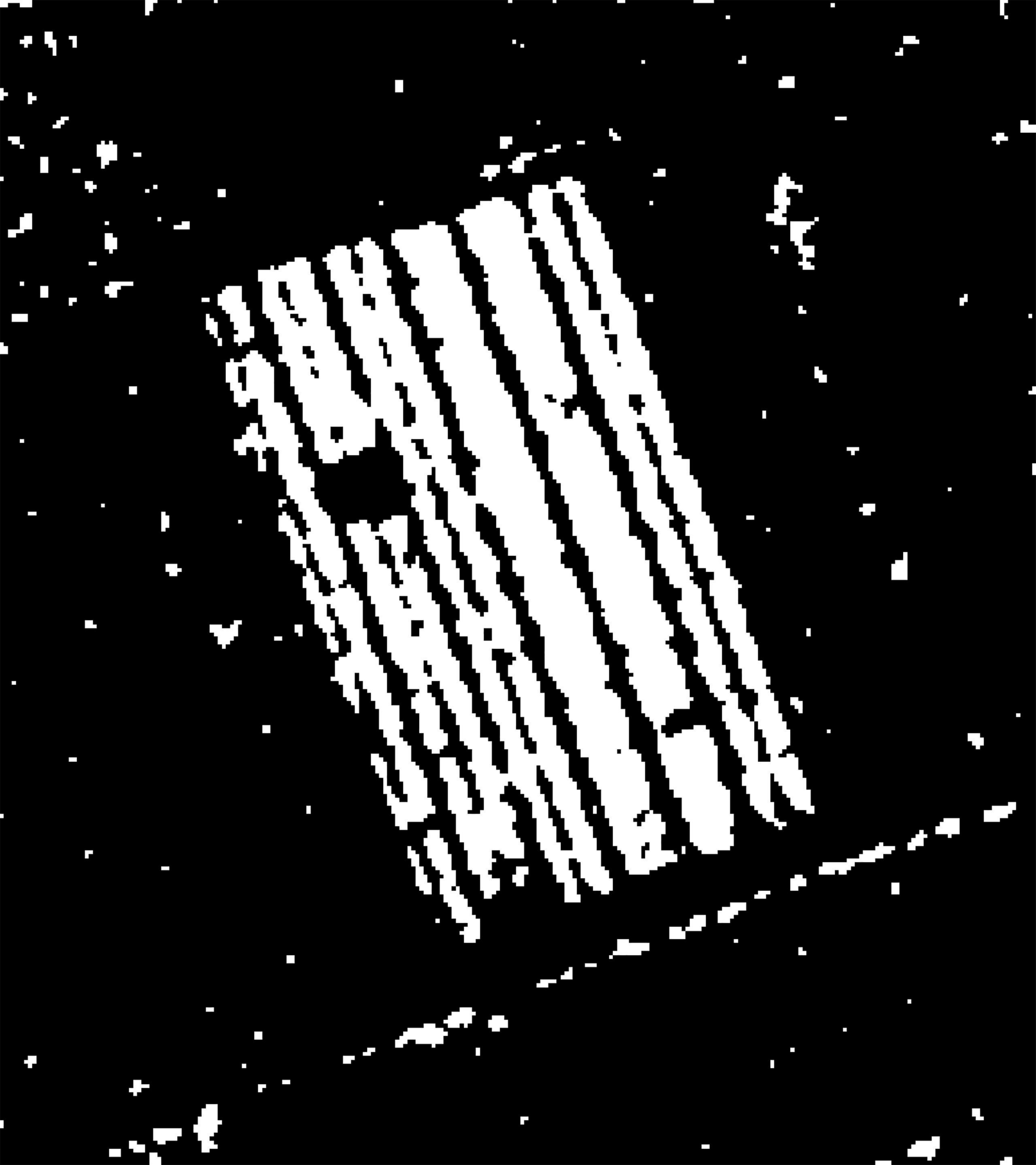}
	}
	\subfigure[]{
		\centering
		\includegraphics[width=0.3\linewidth]{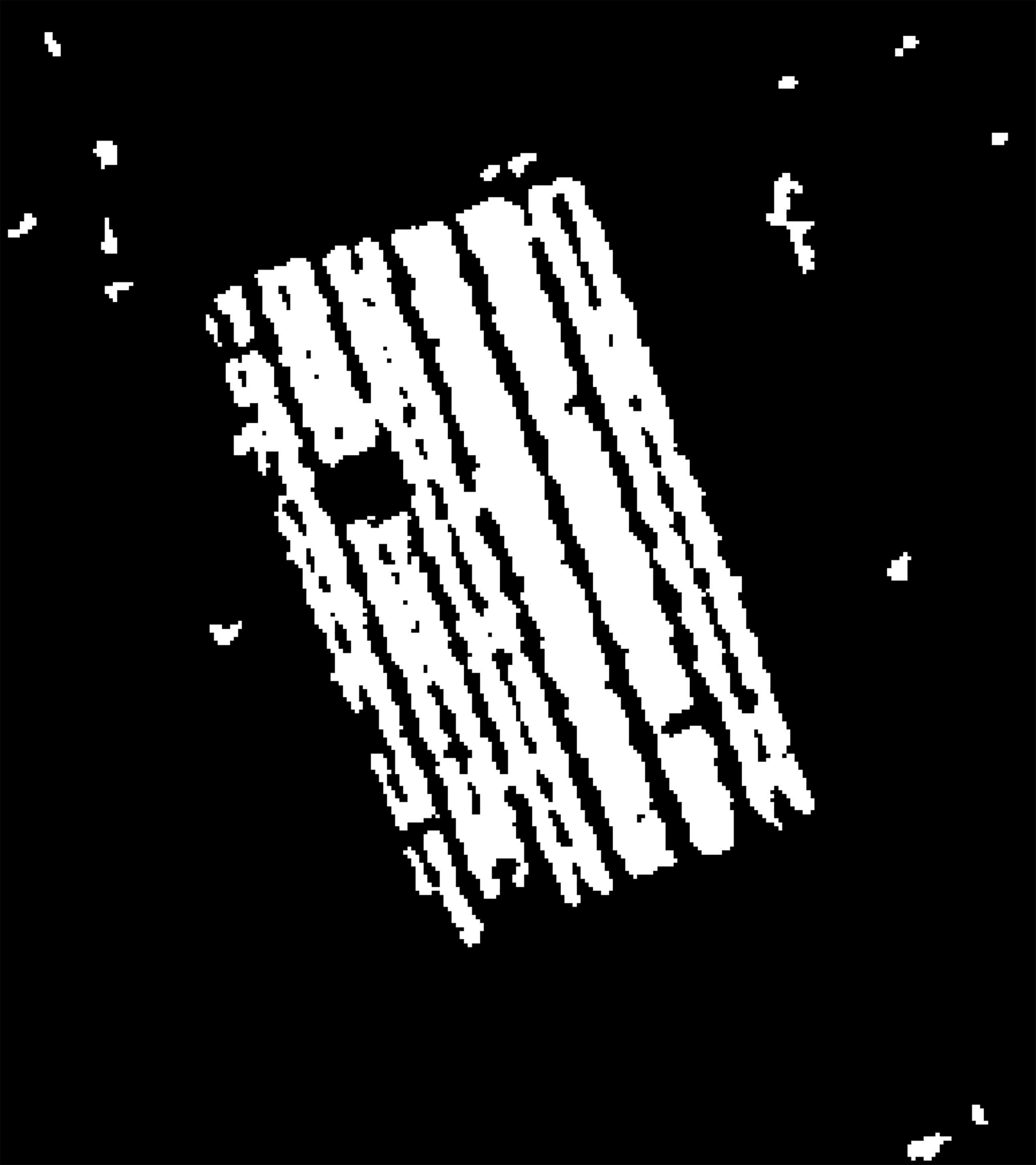}
	}
	
	\caption{Binary maps for Yellow River dataset obtained by different methods. (a) Binary map by LMR. (b) Binary map by PCA. (c) Binary map by PCANet. (d) Binary map by ELM. (e) Binary map by CDML. (f) Binary map by USCNN.}
	\label{figure10}
	
\end{figure}

The final change detection maps on the Yellow River dataset are shown in Fig. \ref{figure10}, and the quantitative evaluation results are presented in Table  \ref{table2}. As shown in Fig.\ref{figure10}(a), (b) and (e), there exists a lot of noise in both images, especially for the LMR. In contrast, PCANet, ELM and USCNN are better at dealing with noise. It should be noted that the white oblique line on the bottom of the change detection image generated by USCNN is completely invisible, which represents the changed region, but is not detected. Therefore, we can see that USCNN produces a lower $Kappa$ coefficient than PCANet from Table III. However, the $PCC$ value is a little larger than PCANet.

\begin{table}[htbp]
	\centering
	\caption{Change detection indicators of different methods on the Yellow River dataset}
	\label{table2}
	\begin{tabular}{cccccc}
		\hline
		Methods & $FP$ & $FN$ & $OE$ & $PCC$ & $Kappa$\\
		\hline
		LMR    &3702  &3212   &6914   &0.9069&	0.6902\\
		PCA	   &1982  &2617	  &4599   &0.9381&	0.7871\\
		PCANet &1741  &1626   &3367	  &0.9547&	\textbf{0.8475}\\
		ELM    &588   &3930   &4518   &0.9391&	0.7726\\
		CDML   &1216  &2223   &3439   &0.9536&	0.8390\\
		USCNN  &1163  &2178   &	3341  &\textbf{0.9550}&	0.8436\\
		\hline
	\end{tabular}
\end{table}

\subsubsection{Result on the  Sardinia dataset}\label{subsubsect3.3.4}

\begin{figure}[htbp]
	\centering
	
	\subfigure[]{
		\centering
		\includegraphics[width=0.3\linewidth]{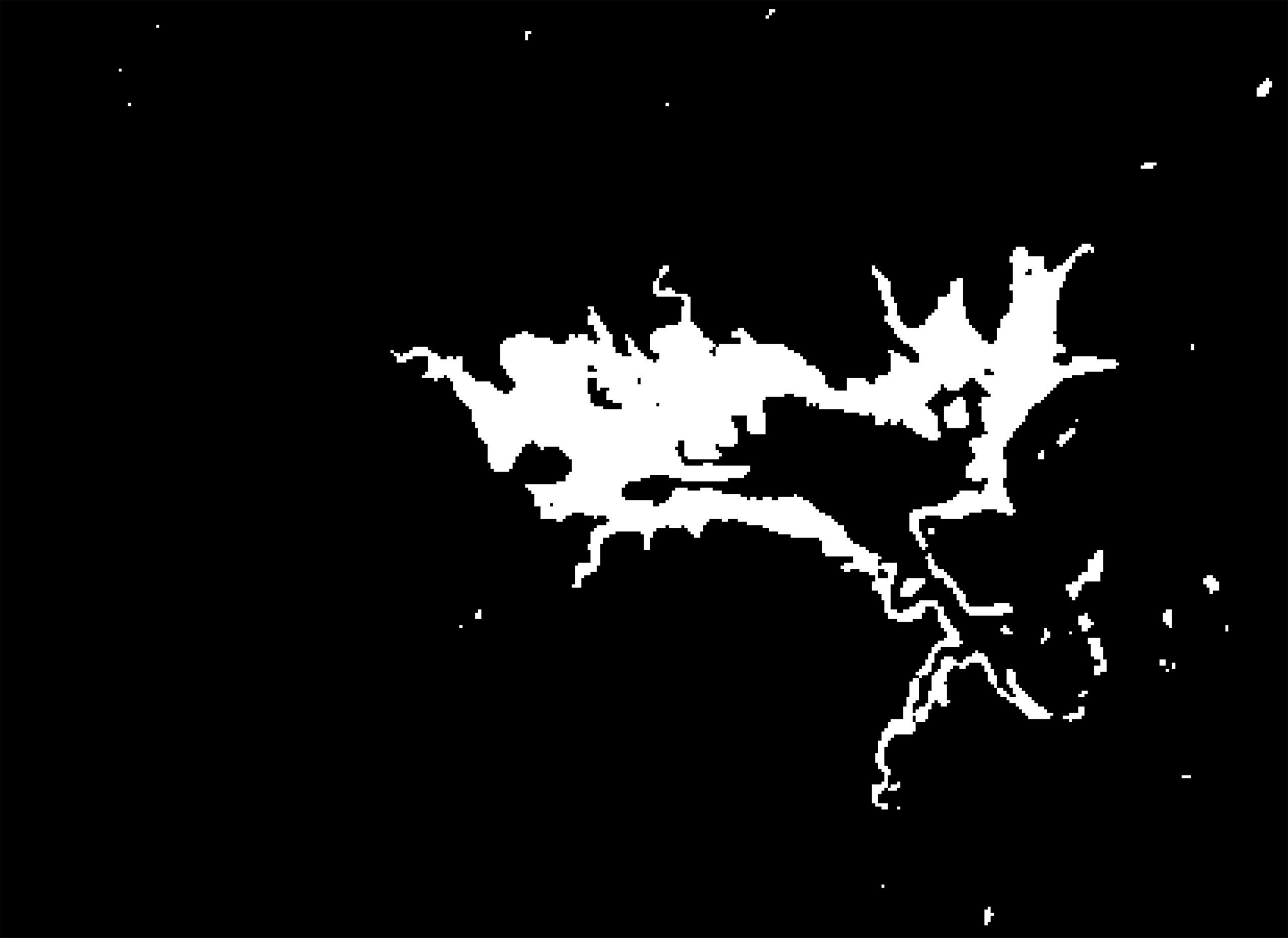}
	}
	\subfigure[]{
		\centering
		\includegraphics[width=0.3\linewidth]{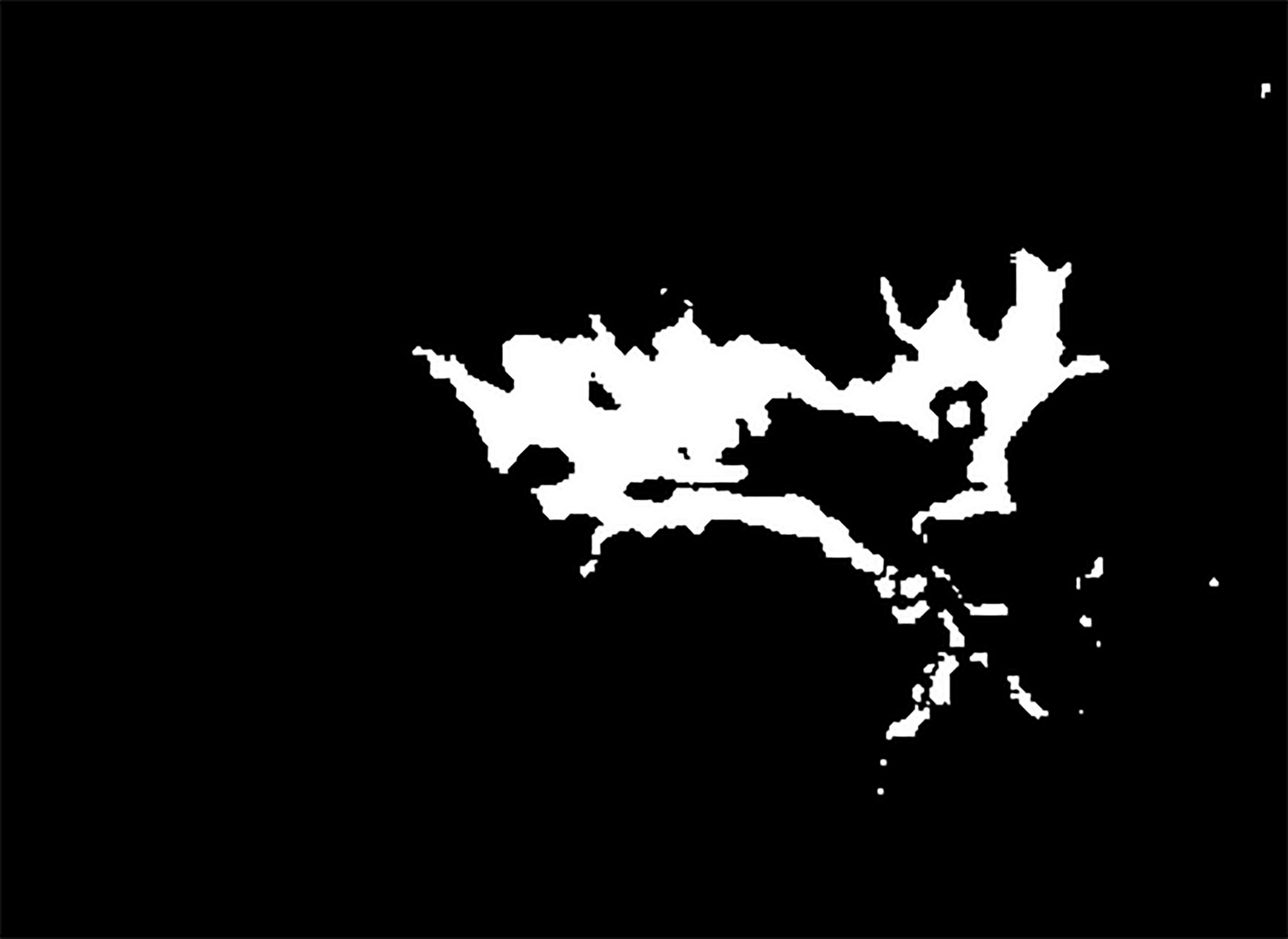}
	}
	\subfigure[]{
		\centering
		\includegraphics[width=0.3\linewidth]{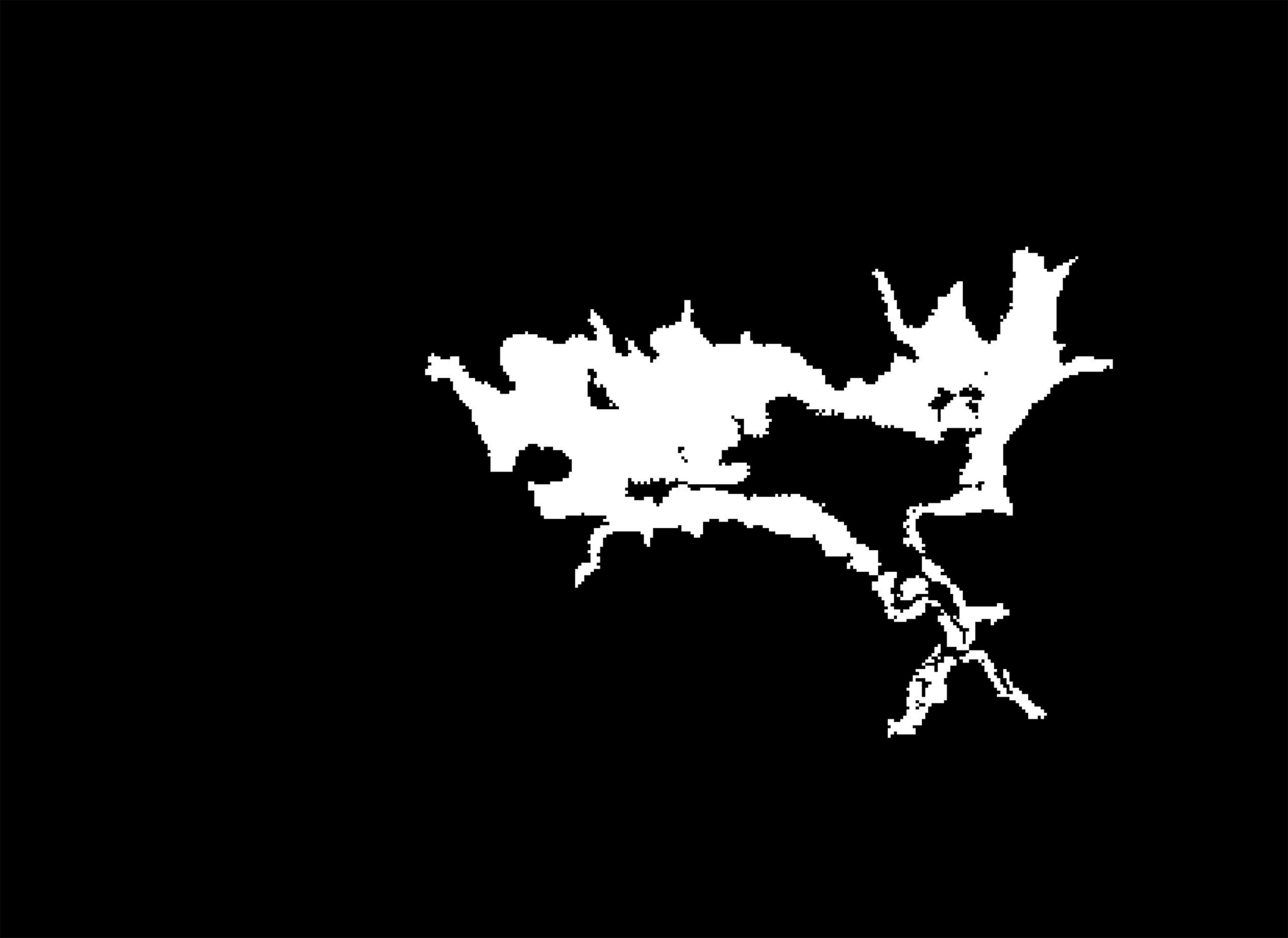}
	}
	\vfill
	\subfigure[]{
		\centering
		\includegraphics[width=0.3\linewidth]{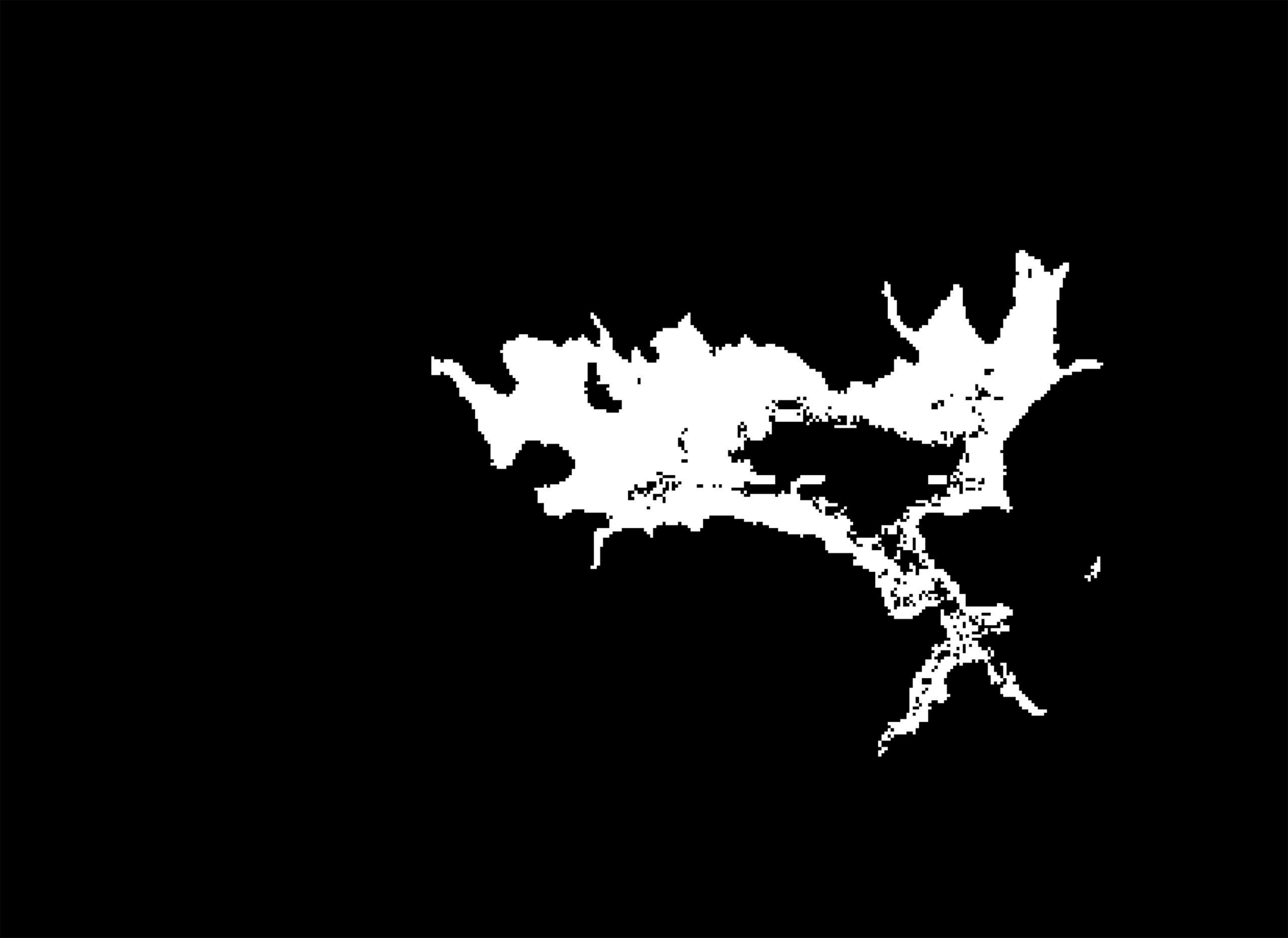}
	}
	\subfigure[]{
		\centering
		\includegraphics[width=0.3\linewidth]{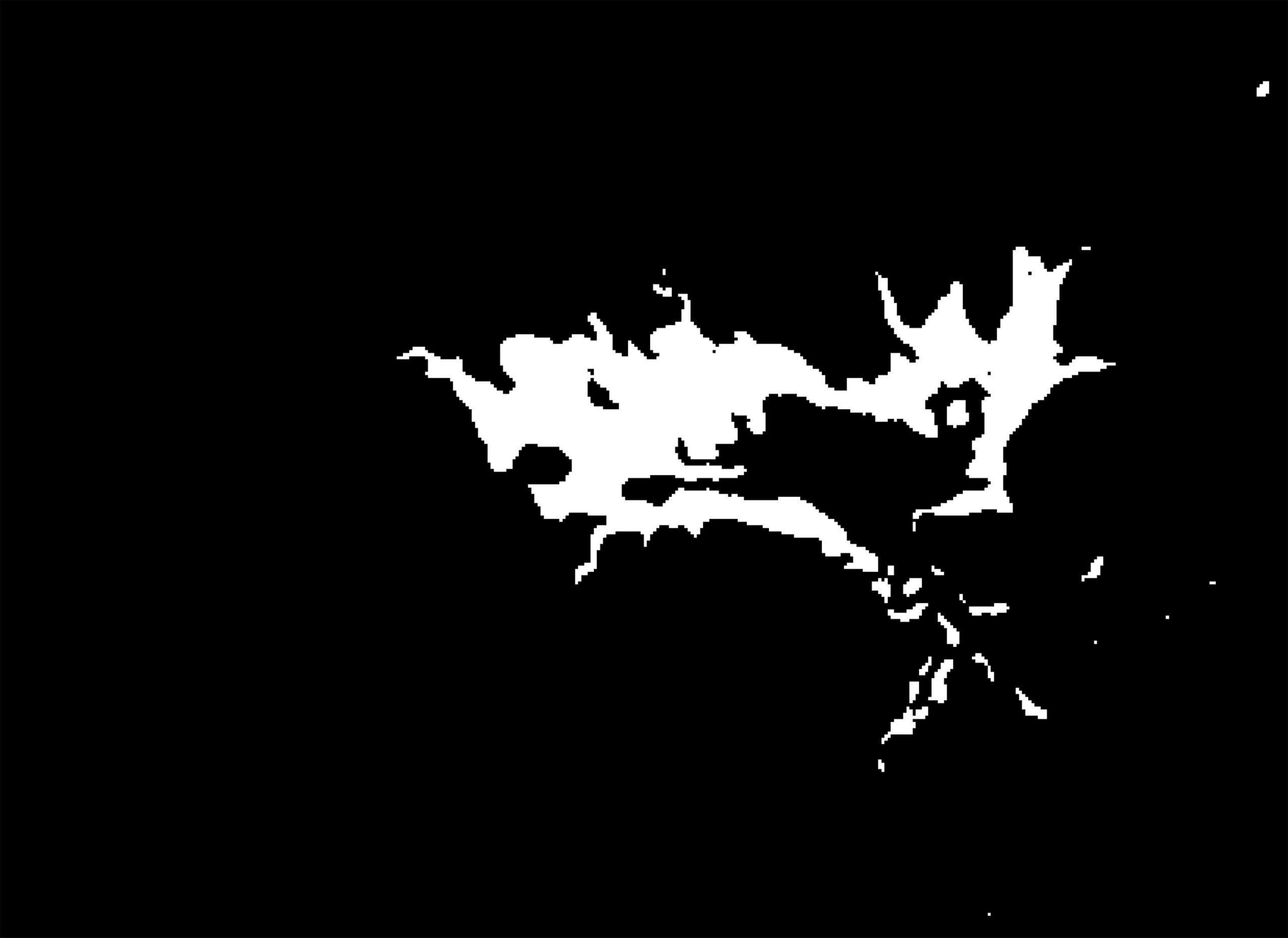}
	}
	\subfigure[]{
		\centering
		\includegraphics[width=0.3\linewidth]{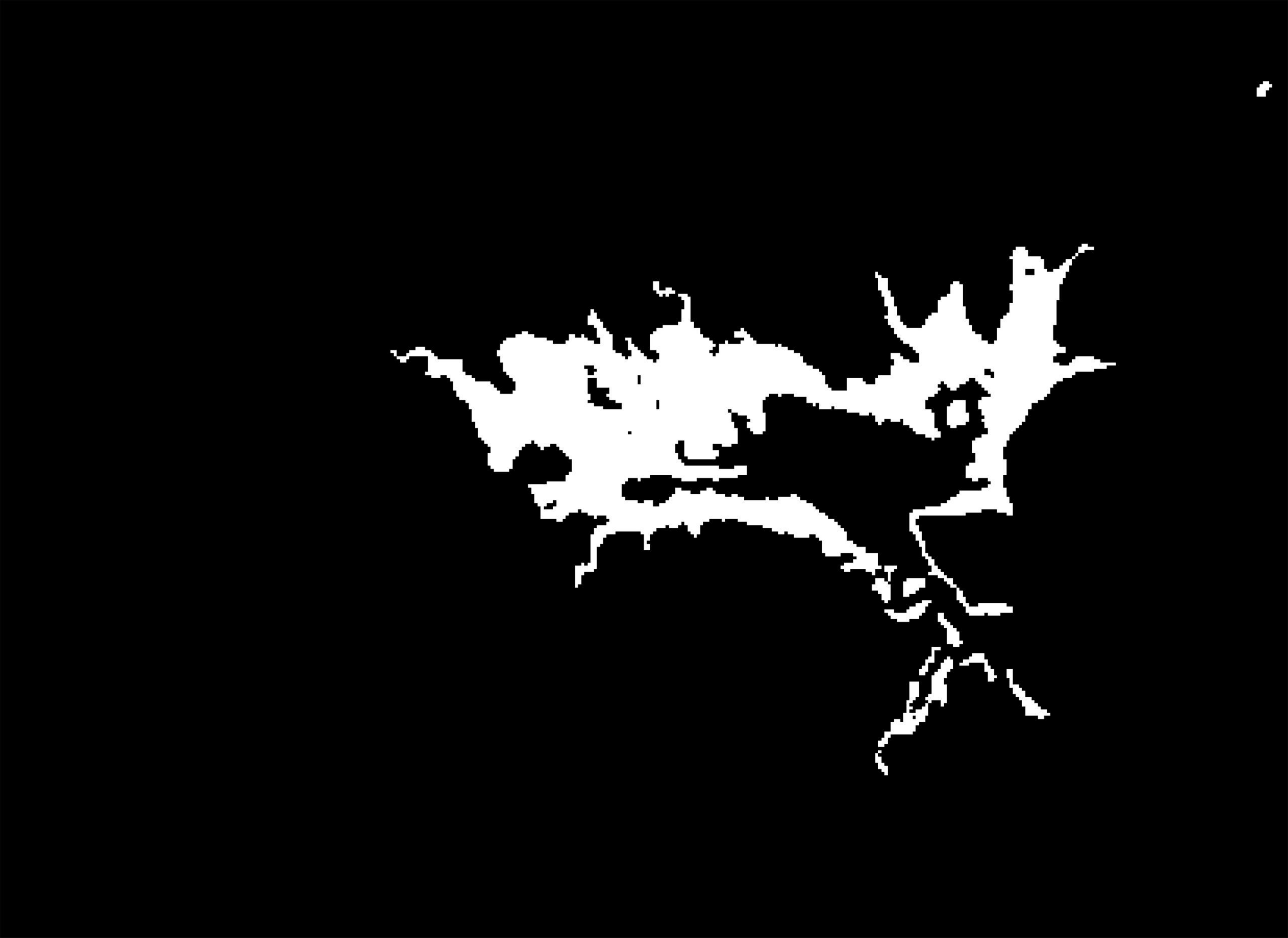}
	}
	
	\caption{Binary maps for Sardinia dataset obtained by different methods. (a) Binary map by LMR. (b) Binary map by PCA. (c) Binary map by PCANet. (d) Binary map by ELM. (e) Binary map by CDML. (f) Binary map by USCNN.}
	\label{figure11}
	
\end{figure}

Fig.\ref{figure11} presents the binary maps generated by LMR, PCA, PCANet, ELM, CDML and USCNN on the Sardinia dataset. The quantitative evaluation results are shown in Table \ref{table3}. As shown in Fig.\ref{figure11}(a), the difference maps by LMR and PCA look noisy. In the result generated by ELM, there are a large number of unchanged pixels that are classified as changed pixels, so it can be seen from Table  \ref{table3} that the $FP$ value is high. It can be seen from Table \ref{table3} that the detection results of our method are very close to those of CDML method, and their PCC are 0.9840 and 0.9842 respectively.

\begin{table}[htbp]
	\centering
	\caption{Change detection indicators of different methods on the Sardinia dataset}
	\label{table3}
	\begin{tabular}{cccccc}
		\hline
		Methods & $FP$ & $FN$ & $OE$ & $PCC$ & $Kappa$\\
		\hline
		LMR    &1970  &524    &2494   &0.9798&	0.8399\\
		PCA	   & 1569 &694    &2263   &0.9817&	0.8499\\
		PCANet &2009  &627    &2636   &0.9787&	0.8301\\
		ELM    &2377  &1025   &3402	  &0.9725&	0.7805\\
		CDML   &1235  &716    &1951   &\textbf{0.9842} &\textbf{0.8678}\\
		USCNN  &1308  &671    &	1979  &0.9840&	0.8669\\
		\hline
	\end{tabular}
\end{table}

\section{Conclusion}\label{sect4}
In this paper, we establish a novel unsupervised convolutional neural network fusion approach for remote sensing change detection. In the proposed method, the idea of the classical log-ratio operator and mean-ratio operator is introduced into the convolutional neural network, and multi-scale filtering kernels are used to extract different scale information from the input images to suppress noise. In addition, an objective function with sparse properties is designed to train the network. The method does not require any pre-detection procedures to provide training samples, thus, the entire training process is performed in an unsupervised manner. The final experimental results have demonstrated the feasibility, robustness and reliability of the proposed method on remote sensing change detection.

\subsection* {Acknowledgments}
This work was supported by the National Natural Science Foundation of China under Grant 61201323 and
Natural Science Foundation projects of Shaanxi Province of China under Grant (2017JM6026, 2018JM6056).


\bibliography{report}   
\bibliographystyle{spiejour}   


\end{spacing}
\end{document}